\documentclass[runningheads]{llncs}

\usepackage{eccv}
\usepackage{eccvabbrv}

\usepackage{graphicx}
\usepackage{booktabs}
\usepackage{amsmath}
\usepackage{amssymb}
\usepackage{multirow}
\usepackage{colortbl}
\usepackage{pifont}
\definecolor{headercolor}{RGB}{218,224,235}
\definecolor{geminicolor}{RGB}{227,237,250}   % soft blue
\definecolor{claudecolor}{RGB}{237,228,248}   % soft lavender
\definecolor{gptcolor}{RGB}{226,243,228}      % soft green
\definecolor{kimicolor}{RGB}{253,237,220}     % soft peach
\definecolor{deepseekcolor}{RGB}{220,243,243} % soft teal
\definecolor{glmcolor}{RGB}{240,235,225}      % soft warm gray
\definecolor{qwencolor}{RGB}{250,228,235}     % soft pink

\usepackage[accsupp]{axessibility}  % Improves PDF readability for those with disabilities.

\usepackage[pagebackref,breaklinks,colorlinks,citecolor=eccvblue]{hyperref}

\usepackage{enumitem}
\usepackage{tabularx}
\usepackage{float}
\usepackage{tcolorbox}
\tcbuselibrary{skins,breakable}
\definecolor{draftbg}{RGB}{255,248,235}
\definecolor{draftframe}{RGB}{230,180,80}
\definecolor{refinedbg}{RGB}{235,248,255}
\definecolor{refinedframe}{RGB}{70,140,210}
\definecolor{genbg}{RGB}{235,250,240}
\definecolor{genframe}{RGB}{60,170,100}
\newtcolorbox{draftplanbox}[1][]{%
  enhanced, breakable,
  colback=draftbg, colframe=draftframe,
  boxrule=0.8pt, arc=4pt,
  fonttitle=\bfseries\small,
  title={#1},
  left=6pt, right=6pt, top=4pt, bottom=4pt,
  before upper={\small},
  before skip=4pt, after skip=4pt,
}

\newtcolorbox{refinedplanbox}[1][]{%
  enhanced, breakable,
  colback=refinedbg, colframe=refinedframe,
  boxrule=0.8pt, arc=4pt,
  fonttitle=\bfseries\small,
  title={#1},
  left=6pt, right=6pt, top=4pt, bottom=4pt,
  before upper={\small},
  before skip=4pt, after skip=4pt,
}

\newtcolorbox{genbox}[1][]{%
  enhanced, breakable,
  colback=genbg, colframe=genframe,
  boxrule=0.8pt, arc=4pt,
  fonttitle=\bfseries\small,
  title={#1},
  left=6pt, right=6pt, top=4pt, bottom=4pt,
  before upper={\small},
  before skip=4pt, after skip=4pt,
}

\definecolor{anticheatingbg}{RGB}{255,240,240}
\definecolor{anticheatingframe}{RGB}{200,60,60}

\newtcolorbox{anticheatingbox}[1][]{%
  enhanced, breakable,
  colback=anticheatingbg, colframe=anticheatingframe,
  boxrule=0.8pt, arc=4pt,
  fonttitle=\bfseries\small,
  title={#1},
  left=6pt, right=6pt, top=4pt, bottom=4pt,
  before upper={\small},
  before skip=4pt, after skip=4pt,
}

\definecolor{validationbg}{RGB}{245,240,255}
\definecolor{validationframe}{RGB}{120,80,200}

\newtcolorbox{validationbox}[1][]{%
  enhanced, breakable,
  colback=validationbg, colframe=validationframe,
  boxrule=0.8pt, arc=4pt,
  fonttitle=\bfseries\small,
  title={#1},
  left=6pt, right=6pt, top=4pt, bottom=4pt,
  before upper={\small},
  before skip=4pt, after skip=4pt,
}
\begin{document}

\title{WebForge: Breaking the Realism-Reproducibility-Scalability Trilemma in Browser Agent Benchmark} 

\titlerunning{WebForge}

% Author list
\author{Peng Yuan\inst{1,2} \and
Yuyang Yin\inst{1} \and
Yuxuan Cai\inst{1} \and
Zheng Wei\inst{1}}

% Abbreviated author list for running head
\authorrunning{P.~Yuan et al.}

% Institution list
\institute{Tencent BAC \and
Tsinghua University}

\maketitle

\begin{abstract}
Existing browser agent benchmarks face a fundamental trilemma: real-website benchmarks lack reproducibility due to content drift, controlled environments sacrifice realism by omitting real-web noise, and both require costly manual curation that limits scalability. We present \textbf{WebForge}, the first fully automated framework that resolves this trilemma through a four-agent pipeline---Plan, Generate, Refine, and Validate---that produces interactive, self-contained web environments end-to-end without human annotation. A seven-dimensional difficulty control framework structures task design along navigation depth, visual complexity, reasoning difficulty, and more, enabling systematic capability profiling beyond single aggregate scores. Using WebForge, we construct \textbf{WebForge-Bench}, a benchmark of 934 tasks spanning 7 domains and 3 difficulty levels. Multi-model experiments show that difficulty stratification effectively differentiates model capabilities, while cross-domain analysis exposes capability biases invisible to aggregate metrics. Together, these results confirm that multi-dimensional evaluation reveals distinct capability profiles that a single aggregate score cannot capture. Code and benchmark will be publicly released.
\keywords{Web Agent Benchmark \and Automated Benchmark Generation \and Multi-dimensional Evaluation \and Browser Agent}
\end{abstract}

\section{Introduction}
\label{sec:intro}

Autonomous web agents that navigate websites, fill forms, and complete multi-step tasks have improved rapidly~\cite{zhou2024webarena,he2024webvoyager,deng2023mind2web,zheng2024seeact,wang2026colorbrowseragent,wei2025webagentr1}. Yet the benchmarks that measure this progress are themselves becoming a bottleneck. We argue that existing approaches cannot simultaneously be \emph{realistic}, \emph{reproducible}, and \emph{scalable}---a tension we term the \textbf{benchmark trilemma}.

\textbf{The benchmark trilemma.}
Real-website benchmarks such as WebVoyager~\cite{he2024webvoyager} maximize realism but suffer from content drift that progressively invalidates tasks---nearly half of Mind2Web's original tasks became outdated within two years~\cite{xue2025onlinemind2web}. Online-Mind2Web~\cite{xue2025onlinemind2web} further exposes how narrow task diversity inflates scores: Browser Use reports {$\sim$}90\% on WebVoyager yet only 30\% on its broader 136-website benchmark with rigorous human evaluation.
Controlled environments such as WebArena~\cite{zhou2024webarena} guarantee reproducibility but remain unrealistically ``clean''---lacking pop-ups, cookie dialogs, and network delays---and depend on extensive manual curation. EntWorld~\cite{mo2026entworld} automates task generation via schema-driven synthesis and SQL-based verification, yet its sandboxed enterprise environment still lacks real-web noise.
Automated generation offers scalability but has not reached the complexity of interactive web environments: BenchAgents~\cite{butt2025benchagents} and AutoBencher~\cite{li2025autobencher} handle only non-interactive tasks; DyVal~\cite{zhu2024dyval} provides four-level complexity control for reasoning tasks via DAG structural parameters, and OS-Genesis~\cite{sun2025osgenesis} produces GUI trajectories with quality-based filtering, but neither generates complete interactive web environments with multi-dimensional difficulty control.

\textbf{WebForge.}
We present WebForge, the first fully automated framework for constructing realistic, reproducible, and scalable browser agent benchmarks with multi-dimensional difficulty control. A four-agent pipeline (Plan $\rightarrow$ Generate $\rightarrow$ Refine $\rightarrow$ Validate) produces interactive web environments end-to-end: the Plan Agent designs tasks with a seven-dimensional difficulty vector $\boldsymbol{\delta} \in \{1,2,3\}^7$; the Generation Agent builds functional websites with real data and anti-cheating mechanisms; the Refinement Agent injects real-web noise (pop-ups, cookie dialogs, network delays); and the Validation Agent verifies solvability. Every environment is a self-contained static website requiring no external services, so deployment reduces to opening an HTML file.

Our contributions are:
\begin{itemize}
\item We identify the benchmark trilemma and resolve it with an end-to-end pipeline that generates complete, interactive web environments with real data, real-web noise, and zero manual annotation.

\item We introduce a seven-dimensional difficulty control framework enabling fine-grained diagnosis of agent capabilities across navigation, interaction, visual, and reasoning axes.

\item We construct WebForge-Bench spanning 7 domains $\times$ 3 difficulty tiers. Multi-model experiments reveal per-dimension capability gaps that aggregate scores obscure.
\end{itemize}

\section{Related Work}
\label{sec:related}

\subsection{Browser Agent Benchmarks}

Browser agent benchmarks span live websites to sandboxed environments, each balancing realism, reproducibility, and scalability differently.

\textbf{Real-website benchmarks.}
Mind2Web~\cite{deng2023mind2web} collected 2{,}350 tasks across 137 websites but evaluates on cached snapshots via step-wise action prediction.
WebVoyager~\cite{he2024webvoyager} tests multimodal agents on 15 live websites; BrowseComp~\cite{wei2025browsecomp} demands deep multi-site browsing (GPT-4o 0.6\%, Deep Research 51.5\%); and GAIA~\cite{mialon2024gaia} requires multi-step reasoning with web tools (human 92\% vs.\ GPT-4 with plugins $\sim$15\%).
However, real-website benchmarks face continuous content drift and silent deprecation---WebCanvas~\cite{pan2024webcanvas} reports that 12\% of its sampled Mind2Web tasks expired within one year. The problem is compounded by evaluation unreliability: BrowserArena~\cite{anupam2025browserarena} finds that GPT-4o as judge agrees with humans only 68\% of the time when evaluating agent traces.

\textbf{Controlled-environment benchmarks.}
WebArena~\cite{zhou2024webarena} hosts four self-contained web applications with 812 tasks and programmatic validators (GPT-4 14.41\%, human 78.24\%).
VisualWebArena~\cite{koh2024visualwebarena} adds 910 visually grounded tasks (GPT-4V+SoM 16.4\%); WebChoreArena~\cite{miyai2025webchorearena} introduces 532 tasks stressing long-term cross-page memory; WorkArena~\cite{drouin2024workarena}/WorkArena++~\cite{drouin2024workarenaplusplus} target ServiceNow workflows (33/682 tasks); and EntWorld~\cite{mo2026entworld} spans six enterprise domains with 1{,}756 tasks and SQL-based verification.
ST-WebAgentBench~\cite{levy2026stwebagentbench} layers safety constraints onto WebArena, where Completion under Policy drops from 24.3\% to 15.0\%.
However, controlled environments remain ``clean'' (no pop-ups, cookie dialogs, or network delays), systematically overstating robustness, and nearly all demand expensive manual curation~\cite{deng2023mind2web,zhou2024webarena,koh2024visualwebarena}. Together, these shortcomings define the \emph{benchmark trilemma}: no existing benchmark achieves high realism, reproducibility, and scalability simultaneously.

\subsection{Automated Benchmark Construction}

Manual benchmark creation is expensive, prompting work on automated alternatives, though none handle interactive environments.
BenchAgents~\cite{butt2025benchagents} automates benchmark creation with a four-agent pipeline but only for non-interactive tasks (calendar scheduling, constrained text generation, and visual causal reasoning).
AutoBencher~\cite{li2025autobencher} optimizes QA-pair quality, producing items 22\% harder than human-curated ones, yet remains limited to static text-based evaluations (QA and safety prompts).
DyVal~\cite{zhu2024dyval}/DyVal2~\cite{zhu2024dyval2} generate controllable-complexity reasoning tasks at the text level; OS-Genesis~\cite{sun2025osgenesis} derives training tasks from GUI exploration but lacks difficulty control and benchmark-oriented evaluation; TaskBench~\cite{shen2024taskbench} synthesizes tasks from tool-use graphs.
None yield interactive web environments with browser-based validation.

\subsection{Multi-dimensional Evaluation}

Several lines of evidence point to the need for multi-dimensional evaluation.
VisualWebBench~\cite{liu2024visualwebbench} finds that its web-specific sub-task rankings correlate poorly with Mind2Web, questioning single-score summaries.
Shlomov~\etal~\cite{shlomov2025grounding2planning} demonstrate that planning accuracy reaches only 86\% even with oracle grounding, concluding that planning, not grounding, is the primary bottleneck.
EntWorld~\cite{mo2026entworld} shows trajectory length as an independent difficulty factor (e.g., GPT-4.1 achieves 34.0\% on short tasks vs.\ 2.7\% on long tasks), while CogAgent~\cite{hong2024cogagent} reveals that performance on general VQA and text-rich VQA benchmarks diverges substantially, a distinction visible only when evaluated separately.
On the agent side, SeeAct~\cite{zheng2024seeact} pinpoints visual grounding as a key bottleneck, and AgentOccam~\cite{yang2025agentoccam} shows that simple well-structured designs can match more complex agents, each highlighting a distinct capability axis.
Despite this evidence, no benchmark offers systematic, a priori difficulty control across multiple dimensions. These observations collectively motivate WebForge's multi-dimensional design.

\section{WebForge: Automated Benchmark Generation Pipeline}
\label{sec:method}

\subsection{Overall Framework}

WebForge constructs browser agent benchmarks through a \textbf{four-stage fully automated pipeline}. Given a target domain $d \in \mathcal{D}$ (7 domains) and difficulty level $l \in \{1, 2, 3\}$:
\begin{equation}
(d, l) \xrightarrow{f_{\text{plan}}} \mathcal{P} \xrightarrow{f_{\text{gen}}} \mathcal{W} \xrightarrow{f_{\text{refine}}} \mathcal{W}^* \xrightarrow{f_{\text{val}}} \{0, 1\}
\end{equation}
where $\mathcal{P}$ is a structured task plan, $\mathcal{W}$ is a web test environment (website files, answer configuration, solution path), and $\mathcal{W}^*$ is the quality-refined environment. Four specialized agents---\textbf{Plan Agent}, \textbf{Generation Agent}, \textbf{Refinement Agent}, and \textbf{Validation Agent}---form a closed loop from task design to quality assurance. A \textbf{seven-dimensional difficulty control framework} governs task design: each dimension (\eg, navigation depth, visual complexity, reasoning difficulty) is independently set at three levels, with the overall difficulty imposing combinatorial constraints on per-dimension levels.

\begin{figure}[t]
\centering
\includegraphics[width=\linewidth]{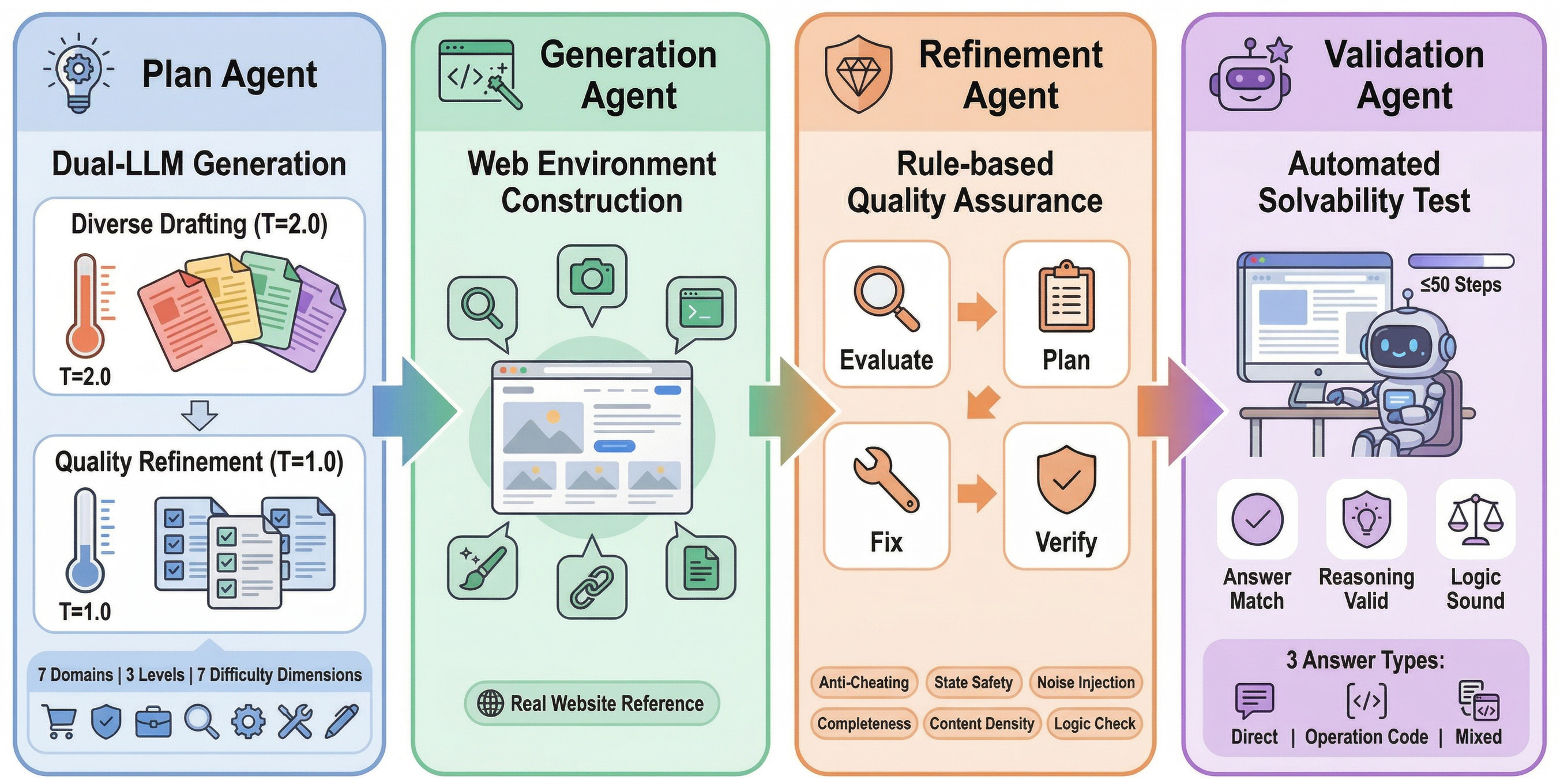}
\caption{\textbf{Overview of the WebForge four-stage automated generation pipeline.} The pipeline is executed sequentially by Plan Agent (blue), Generation Agent (green), Refinement Agent (orange), and Validation Agent (purple).}
\label{fig:overall_pipeline}
\end{figure}

\subsection{Plan Agent: Dual-Stage Task Planning}
\label{sec:plan_agent}

\begin{figure}[t]
\centering
\includegraphics[width=\linewidth]{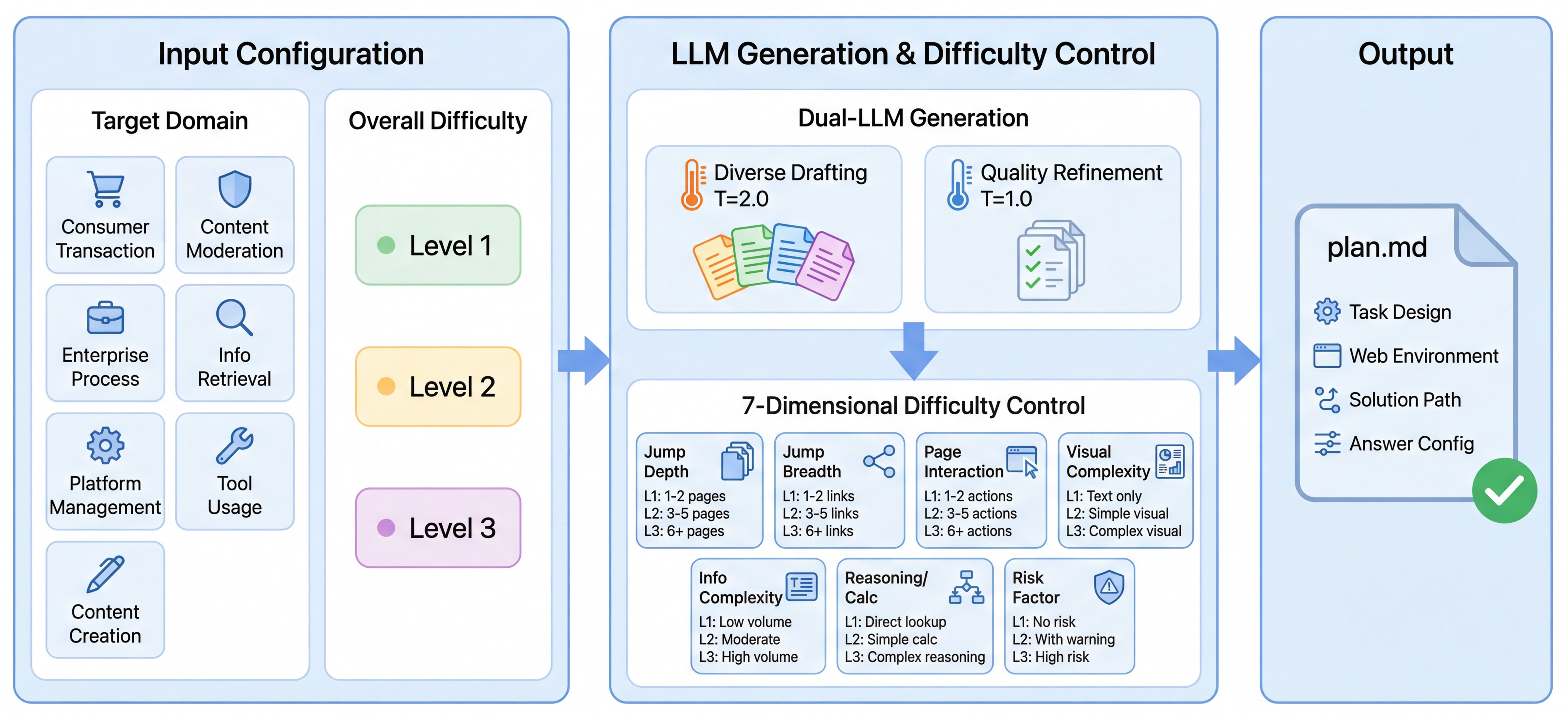}
\caption{\textbf{Plan Agent workflow.} A dual-LLM generation process: high-temperature drafting ($T_h = 2.0$) produces diverse creative proposals, followed by low-temperature refinement ($T_l = 1.0$) for constraint verification and quality enhancement.}
\label{fig:plan_agent}
\end{figure}

Given a target domain $d$ and difficulty level $l$, Plan Agent converts $(d, l)$ into a structured task blueprint $\mathcal{P}$:
\begin{equation}
\mathcal{P} = f_{\text{refine}}^{\text{LLM}}(f_{\text{draft}}^{\text{LLM}}(d, l;\, T_h);\, T_l)
\end{equation}
The two stages employ \emph{different} LLMs. In \textbf{Stage~1} (creative divergence), a creativity-oriented model drafts the task at high temperature $T_h = 2.0$, producing the task objective, seven-dimensional difficulty configuration, web environment design, solution path, and answer type. In \textbf{Stage~2} (quality refinement), a precision-oriented model reviews the draft at low temperature $T_l = 1.0$, performing logic verification, quality enhancement, and difficulty rule validation, modifying at least 30--50\% of the draft. The output blueprint $\mathcal{P}$ includes a difficulty vector $\boldsymbol{\delta} \in \{1,2,3\}^7$ subject to $C(l, \boldsymbol{\delta}) = \texttt{true}$.

\subsection{Generation Agent: Realistic Web Environment Construction}
\label{sec:gen_agent}

Given the plan $\mathcal{P}$ from Plan Agent, Generation Agent instantiates it into a fully runnable web test environment:
\begin{equation}
\mathcal{W} = f_{\text{gen}}(\mathcal{P}) = (\mathcal{S}, \mathcal{A}, \mathcal{M})
\end{equation}
where $\mathcal{S}$ is the set of website files (HTML/CSS/JS pages, image assets, encrypted data files), $\mathcal{A}$ is the answer configuration (task instruction, ground truth, and solution steps), and $\mathcal{M}$ is the website metadata (page structure, navigation graph, asset statistics).

\begin{figure}[t]
\centering
\includegraphics[width=\linewidth]{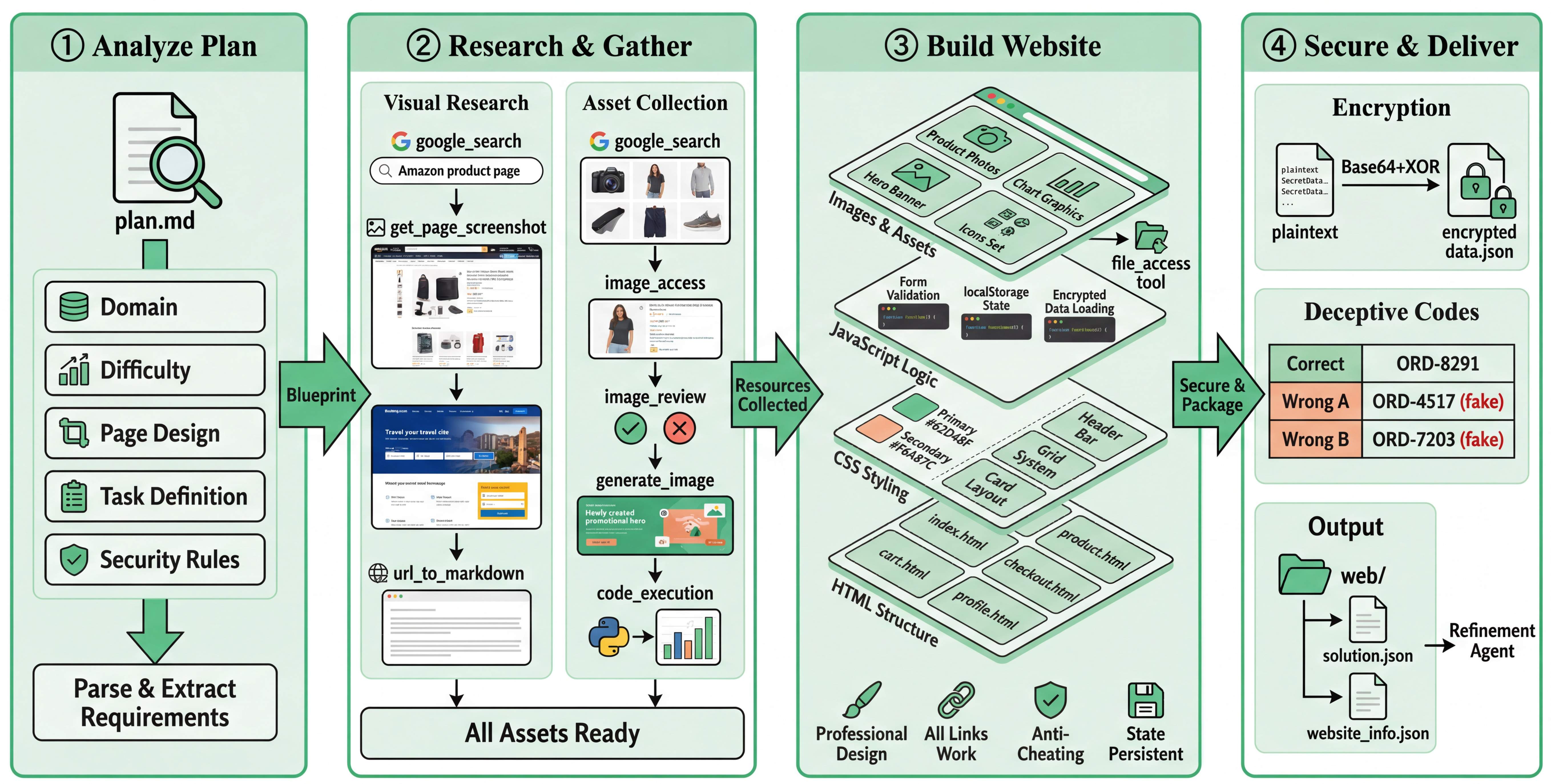}
\caption{\textbf{Generation Agent workflow.} From analyzing the plan blueprint through resource collection to building a complete website with anti-cheating mechanisms.}
\label{fig:gen_agent}
\end{figure}

\noindent\textbf{Workflow.} The agent completes construction through four stages:

\noindent\textit{1.~Analyze plan}---Interpret $\mathcal{P}$ and flexibly adjust details (\eg, replacing placeholders with real data via search) while preserving difficulty levels and domain intent.

\noindent\textit{2.~Resource collection}---Search real websites for visual design references and collect real-world data (images, product information, news text) to embed.

\noindent\textit{3.~Website construction}---Every navigation element links to a real page; \texttt{localStorage}-based state management enables stateful interactions (\eg, shopping carts, form submissions) within purely static pages; the website must not hint at the task objective.

\noindent\textit{4.~Secure and deliver}---WebForge adopts a \emph{final-state evaluation paradigm}: evaluation only checks whether the agent's final output matches the ground truth, granting maximum freedom in path exploration. Three answer types are supported: (i)~\textbf{Direct Answer}---the agent reports a concrete value (\eg, a price); (ii)~\textbf{Operation Code}---the website embeds a self-contained judging mechanism computing a unique code from the agent's accumulated state; and (iii)~\textbf{Mixed}. In all cases, the evaluator LLM performs a straightforward comparison, eliminating complex semantic judgment. Anti-cheating mechanisms include encrypted data storage, deceptive error codes, and code obfuscation.

\subsection{Refinement Agent: Rule-Driven Quality Assurance}
\label{sec:refine_agent}

Refinement Agent performs systematic inspection and enhancement of the Generation Agent's output:
\begin{equation}
\mathcal{W}^* = f_{\text{refine}}(\mathcal{W}, \mathcal{R})
\end{equation}
where $\mathcal{R}$ is a comprehensive set of quality rules covering functional completeness, visual correctness, state determinism, environment realism, task security, and interaction feedback.

\begin{figure}[t]
\centering
\includegraphics[width=\linewidth]{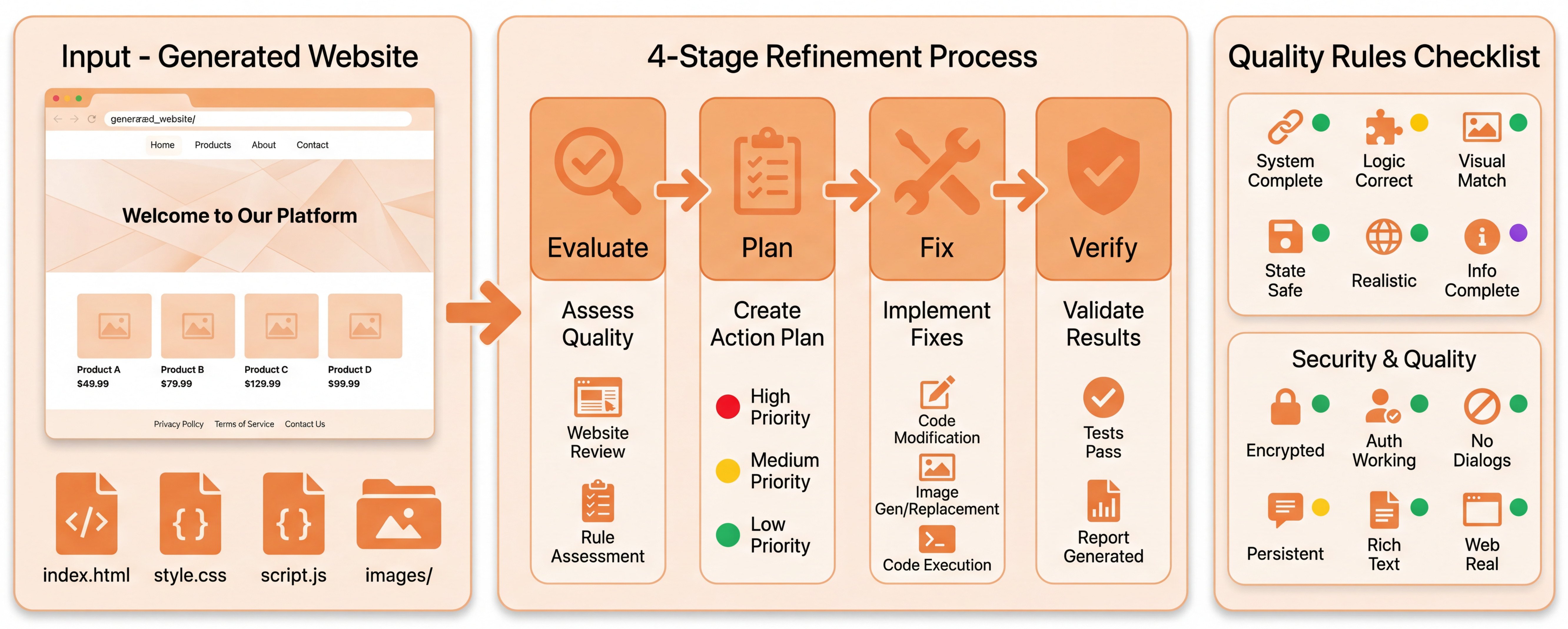}
\caption{\textbf{Refinement Agent workflow.} A four-stage process---Assess Quality, Plan Repairs, Execute Improvements, Verify \& Deliver---guided by a comprehensive quality rules checklist.}
\label{fig:refine_agent}
\end{figure}

\noindent\textbf{Workflow.} The agent follows an \textit{Assess $\rightarrow$ Plan $\rightarrow$ Execute $\rightarrow$ Verify} workflow (\cref{fig:refine_agent}): evaluate the website against the rules checklist, formulate a priority-ranked repair plan, execute modifications, and verify each fix. Among the rules, \textbf{real-web noise injection} is key for environment realism---it injects pop-up advertisements, cookie consent prompts, network latency simulation, and other real-web disturbances, bridging the gap between ``sterile'' virtual environments and the real web.

\subsection{Validation Agent and Evaluation Design}
\label{sec:val_agent}

\begin{figure}[t]
\centering
\includegraphics[width=\linewidth]{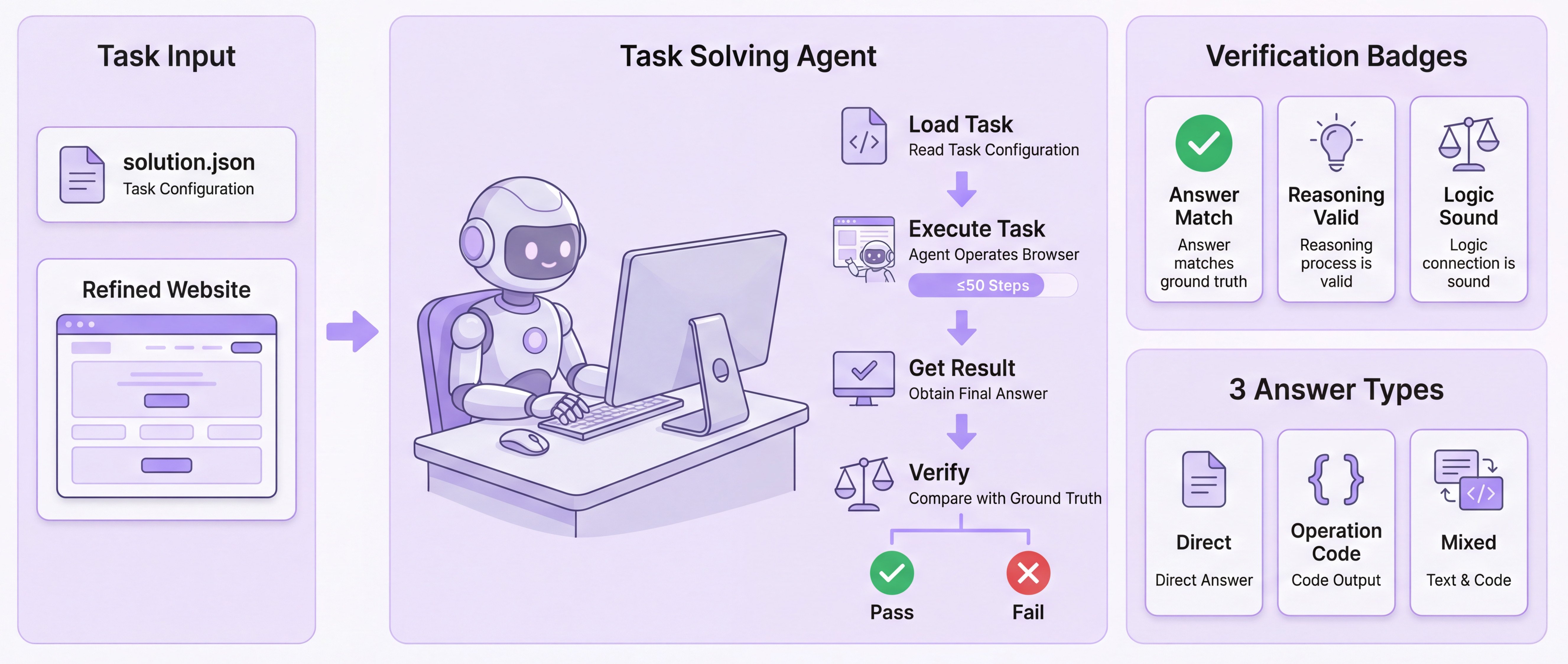}
\caption{\textbf{Validation Agent workflow.} The agent reads the solution file, replays browser actions in an Observe--Reason--Act loop (up to 50 steps), and produces a solvability verdict by comparing the final result against ground truth. Three failure modes are detected: ground-truth mismatch, reasoning logic flaws, and repeated action failures.}
\label{fig:val_agent}
\end{figure}

Validation Agent is the final quality gate of the pipeline, providing a machine-verified certificate that every retained task is solvable within a bounded number of browser actions. Crucially, it operates inside the same Chromium-based browser engine that tested agents will use at evaluation time, unlike the preceding Refinement Agent which works at the source-code level without rendering. This allows it to detect rendering-dependent issues, JavaScript execution errors, and dynamic interaction failures invisible in the source code. Generation and Refinement Agents can produce tasks whose solution paths contain logical errors (\eg, a step references a UI element that does not exist after rendering) or whose ground truth is unreachable due to subtle runtime bugs. By actually executing the solution in the same browser environment that agents will face, the Validation Agent catches these issues and filters out defective tasks.

Given $\mathcal{W}^*$ and its solution path $(a_1, \ldots, a_T)$, the agent replays every action in a real browser:
\begin{equation}
f_{\text{val}}(\mathcal{W}^*) =
\begin{cases}
1 & \text{if } s_T = s^* \text{ and } T \leq 50, \\
0 & \text{otherwise.}
\end{cases}
\end{equation}

\noindent\textbf{Workflow.} The agent operates in an Observe--Reason--Act loop~\cite{yao2023react}: at each step it captures a screenshot and the DOM tree to observe the current page state, reasons about the next action according to the solution path, and executes a browser action. Three checkpoints collectively determine the verdict: (1)~verify that the solution path's reasoning logic correctly derives $s^*$; (2)~replay browser actions step by step, with a 50-step upper bound; and (3)~strictly compare the final result against $s^*$. A 3-retry mechanism handles transient failures---if the same action fails three consecutive times, the task is flagged as unsolvable. Only tasks passing all checks enter the final benchmark; failed tasks are routed back for repair or discarded.

\section{Experiments}
\label{sec:experiments}

\subsection{Experimental Setup}

\noindent\textbf{WebForge-Bench.} We use the WebForge pipeline to generate 1{,}260 tasks (60 per domain--level pair across 7 domains $\times$ 3 difficulty levels). After Validation Agent filtering, 934 tasks pass solvability verification, yielding an overall pipeline pass rate of 74.1\%.

\noindent\textbf{Evaluated models.} We evaluate 14 model configurations spanning closed-source and open-source categories.
\textit{Closed-source multimodal:}
Gemini-3-Pro~\cite{gemini3pro2025},
Gemini-3-Flash~\cite{gemini3flash2025}, and
Gemini-2.5-Flash-Lite~\cite{google2025gemini25flashlite};
Claude-4.5-Sonnet~\cite{anthropic2025claude45sonnet};
GPT-5.2~\cite{openai2025gpt52},
GPT-5-Mini~\cite{openai2025gpt5mini}, and
GPT-5-Nano~\cite{openai2025gpt5nano}.
\textit{Open-source multimodal:}
Kimi-K2.5~\cite{moonshotai2026kimik25};
Qwen3-VL-235B~\cite{qwen2025qwen3vl} and
Qwen3-Omni-30B~\cite{qwen2025qwen3omni}.
\textit{Text-only:}
DeepSeek-V3.2~\cite{deepseekai2025deepseekv32} and
GLM-4.7~\cite{zhipuai2025glm47}.

\noindent\textbf{Evaluation protocol.} All evaluations are conducted in a Chromium-based browser in GUI (non-headless) mode, matching the environment used by the Validation Agent. Each task allows up to 50 browser actions. WebForge adopts a final-state paradigm: it does not monitor intermediate steps but only checks whether the tested agent's output matches the ground truth, granting maximum freedom in path exploration. Three answer types are supported---Direct Answer (exact value matching), Operation Code (the anti-cheating mechanism described in \cref{sec:gen_agent}), and Mixed. In all cases, an evaluator LLM performs a straightforward comparison between the agent's output and the ground truth, eliminating the need for complex semantic judgment or human annotation.

\subsection{Main Results}

\cref{tab:main_results} presents the accuracy of all models across three difficulty levels, grouped by input modality.

\begin{table*}[t]
\centering
\scriptsize
\setlength{\tabcolsep}{3.5pt}
\caption{\textbf{Main results on WebForge-Bench} (934 tasks). Left: accuracy (\%) by difficulty level. Right: cross-domain accuracy (\%). D1: Consumer Transaction/Service, D2: Content Moderation/Compliance, D3: Enterprise Process/Collaboration, D4: Info Retrieval/Analysis, D5: Platform Management/Ops, D6: Tool Usage, D7: Content Creation/Publishing. (a)~Multimodal models receive screenshot + DOM. (b)~Text-only models receive DOM only; rows marked (T) denote multimodal models evaluated with DOM only.}
\label{tab:main_results}
\label{tab:domain}
\begin{tabular}{@{}lcccc!{\vrule width 1.2pt}ccccccc@{}}
\toprule
\rowcolor{headercolor}
 & \multicolumn{4}{c!{\vrule width 1.2pt}}{\textbf{Difficulty Level}} & \multicolumn{7}{c}{\textbf{Cross-Domain}} \\
\rowcolor{headercolor}
\textbf{Model} & \textbf{L1} & \textbf{L2} & \textbf{L3} & \textbf{ALL} & \textbf{D1} & \textbf{D2} & \textbf{D3} & \textbf{D4} & \textbf{D5} & \textbf{D6} & \textbf{D7} \\
\midrule
\multicolumn{12}{@{}l}{\textit{(a) Multimodal (Screenshot + DOM)}} \\
\multicolumn{12}{@{}l}{\textit{\quad Closed-source}} \\
\rowcolor{geminicolor}
Gemini-3-Pro & \textbf{86.4} & \textbf{82.1} & \textbf{58.0} & \textbf{75.9} & \textbf{72.2} & \underline{67.2} & \textbf{82.4} & \textbf{79.4} & \underline{71.0} & \textbf{76.6} & \textbf{80.9} \\
\rowcolor{geminicolor}
Gemini-3-Flash & 82.4 & 73.5 & 44.0 & 67.1 & \underline{65.2} & 61.6 & 66.4 & 62.5 & \textbf{74.0} & 66.0 & 74.8 \\
\rowcolor{geminicolor}
Gemini-2.5-Flash-Lite & 58.5 & 33.5 & 12.6 & 35.0 & 34.8 & 28.8 & 26.7 & 41.9 & 38.2 & 33.3 & 39.7 \\
\rowcolor{claudecolor}
Claude-4.5-Sonnet & \underline{85.7} & \underline{74.7} & \underline{48.1} & \underline{69.9} & 58.3 & \textbf{70.4} & \underline{71.8} & 73.8 & 69.5 & \underline{67.4} & \underline{76.3} \\
\rowcolor{gptcolor}
GPT-5.2 & 80.1 & 65.9 & 31.1 & 59.5 & 48.7 & 58.4 & 51.1 & 64.4 & 57.3 & 63.1 & 71.0 \\
\rowcolor{gptcolor}
GPT-5-Mini & 82.4 & 68.2 & 28.7 & 60.4 & 51.3 & 56.8 & 50.4 & 73.8 & 60.3 & 58.2 & 67.9 \\
\rowcolor{gptcolor}
GPT-5-Nano & 61.8 & 25.9 & 6.1 & 31.3 & 20.9 & 29.6 & 29.0 & 43.8 & 31.3 & 29.8 & 30.5 \\
\multicolumn{12}{@{}l}{\textit{\quad Open-source}} \\
\rowcolor{kimicolor}
Kimi-K2.5 & 84.4 & 73.8 & 39.2 & 66.4 & 60.0 & 61.6 & 65.6 & \underline{75.6} & 62.6 & 61.7 & 74.8 \\
\rowcolor{qwencolor}
Qwen3-VL-235B & 73.4 & 50.3 & 20.1 & 48.3 & 37.4 & 40.8 & 46.6 & 58.8 & 51.1 & 48.2 & 51.1 \\
\rowcolor{qwencolor}
Qwen3-Omni-30B & 26.9 & 9.1 & 2.4 & 12.7 & 6.1 & 9.6 & 7.6 & 26.2 & 10.7 & 12.1 & 13.0 \\
\midrule
\multicolumn{12}{@{}l}{\textit{(b) Text-only (DOM only)}} \\
\multicolumn{12}{@{}l}{\textit{\quad Text-only models}} \\
\rowcolor{deepseekcolor}
DeepSeek-V3.2 & 77.1 & 47.4 & 21.5 & 48.8 & 54.8 & 46.4 & 48.9 & 45.6 & 49.6 & 48.2 & 49.6 \\
\rowcolor{glmcolor}
GLM-4.7 & 76.4 & 49.4 & 24.2 & 50.2 & 50.4 & 43.2 & 55.7 & 48.8 & 52.7 & 48.9 & 51.9 \\
\multicolumn{12}{@{}l}{\textit{\quad Multimodal models (DOM only)}} \\
\rowcolor{geminicolor}
Gemini-3-Pro (T) & 80.1 & 61.8 & 34.8 & 59.2 & 61.7 & 56.0 & 61.1 & 57.5 & 59.5 & 56.7 & 62.6 \\
\rowcolor{geminicolor}
Gemini-3-Flash (T) & 78.7 & 50.9 & 23.2 & 51.2 & 54.8 & 45.6 & 52.7 & 43.8 & 55.0 & 51.8 & 56.5 \\
\midrule
\rowcolor{headercolor}
\textbf{Average} & 73.9 & 54.8 & 28.1 & 52.6 & 48.3 & 48.3 & 51.1 & 56.9 & 53.1 & 51.6 & 57.2 \\
\bottomrule
\end{tabular}
\end{table*}

\noindent\textbf{Difficulty levels provide strong stratification.} The three-tier difficulty system produces clear performance separation. On Level~1 tasks, most models achieve $\geq$73\%, confirming that these tasks are appropriately accessible, while smaller models already begin to struggle (Qwen3-Omni-30B: 26.9\%). Performance drops substantially at Level~2 (9--82\%) and Level~3 becomes highly discriminative (2--58\%), with a 56-point gap between the strongest (Gemini-3-Pro, 58.0\%) and weakest (Qwen3-Omni-30B, 2.4\%) models. This progressive difficulty gradient validates the effectiveness of our seven-dimensional difficulty control framework (see \cref{tab:per_dimension} for a per-dimension breakdown).

\noindent\textbf{Closed-source frontier models lead overall.}
Gemini-3-Pro achieves the highest overall accuracy (75.9\%), followed by Claude-4.5-Sonnet (69.9\%) and Gemini-3-Flash (67.1\%). Among open-source models, Kimi-K2.5 (66.4\%) is the strongest, surpassing closed-source GPT-5.2 (59.5\%).

The right half of \cref{tab:main_results} reveals pronounced cross-domain performance variation that aggregate scores mask. The average row shows that D4 (Info Retrieval/Analysis, 56.9\%) and D7 (Content Creation/Publishing, 57.2\%) are the easiest domains, while D1 (Consumer Transaction/Service, 48.3\%) and D2 (Content Moderation/Compliance, 48.3\%) are the hardest, a gap of nearly 9 points.

\noindent\textbf{Info Retrieval (D4) is the easiest domain.} D4 ranks as the best-performing domain for 6 of 14 models, with an average accuracy of 56.9\%. This aligns with the core strength of current LLMs: information retrieval tasks predominantly require locating and extracting factual content from structured or semi-structured pages---a pattern that closely mirrors pre-training data distributions and retrieval-augmented generation paradigms. The answers tend to be deterministic and verifiable (e.g., finding a specific price, date, or statistic), reducing the need for multi-step reasoning or stateful interaction. Similarly, D7 (Content Creation, 57.2\%) benefits from models' strong text generation capabilities, where producing or editing content aligns naturally with language modeling objectives.

\noindent\textbf{Consumer Transaction (D1) and Content Moderation (D2) are universally hard.} D1 or D2 appears as the worst domain for 9 of 14 models, both averaging only 48.3\%. Consumer Transaction tasks demand complex \emph{stateful multi-step workflows}---adding items to carts, filling shipping forms, applying coupons, and completing checkout sequences---where a single misstep in any intermediate state can cascade into task failure. These tasks also involve \emph{irreversible operations} (e.g., confirming a purchase) that require careful planning before execution. Content Moderation tasks pose a different challenge: they require \emph{policy-grounded judgment} over nuanced content (e.g., distinguishing borderline violations from acceptable posts), a capability that current LLMs, trained primarily on general-purpose corpora, have limited exposure to. The GPT series illustrates this domain sensitivity most starkly: GPT-5-Mini leads on D4 (73.8\%) yet drops to 50.4\% on D3, a gap exceeding 23 points. These domain-specific biases are invisible in single-domain benchmarks, underscoring the value of cross-domain evaluation.

\subsection{Ablation Studies}

\noindent\textbf{Effect of visual input.}
\cref{tab:main_results}(b) directly reveals the impact of visual input. Comparing multimodal models in part~(a) with their text-only variants in part~(b), removing screenshots causes consistent accuracy drops of 16--17 percentage points overall (Gemini-3-Pro: 75.9\%$\rightarrow$59.2\%; Gemini-3-Flash: 67.1\%$\rightarrow$51.2\%), confirming that WebForge tasks genuinely require visual understanding. The gap widens with difficulty: $\sim$6 points at Level~1, 16--23 at Level~2, and 20+ at Level~3, aligning with greater visual complexity by design. Notably, the text-only models DeepSeek-V3.2 (48.8\%) and GLM-4.7 (50.2\%) perform comparably to the text-only variants of multimodal models (51--59\%), suggesting that their moderate rankings are partly attributable to the absence of visual information rather than weaker reasoning.

\noindent\textbf{Pipeline component ablation.}
To validate the contribution of each pipeline stage, we conduct ablation experiments on a 210-task subset (10 per domain--level pair). \cref{tab:ablation_pipeline} reports the Validation Agent pass rate under three configurations. The full pipeline achieves 74.1\% (934/1{,}260). Removing Plan Agent's refinement stage (Stage~2, \cref{sec:plan_agent}) drops the rate to 59.5\%, because without refinement the generated plans lack sufficient feasibility and logical rigor, leading to tasks that the Generation Agent cannot faithfully implement. Further removing Refinement Agent reduces the rate to 51.4\%, demonstrating that each stage contributes meaningfully to task quality.

\begin{table}[t]
\centering
\small
\caption{\textbf{Pipeline ablation.} Validation pass rate under different pipeline configurations.}
\label{tab:ablation_pipeline}
\begin{tabular}{@{}ccc@{}}
\toprule
\rowcolor{headercolor}
\textbf{Plan Refine} & \textbf{Refine Agent} & \textbf{Pass Rate (\%)} \\
\midrule
\ding{55} & \ding{55} & 51.4 \\
\ding{55} & \ding{51} & 59.5 \\
\ding{51} & \ding{51} & \textbf{74.1} \\
\bottomrule
\end{tabular}
\end{table}

\subsection{Runtime Efficiency Analysis}

\begin{table*}[t]
\centering
\scriptsize
\setlength{\tabcolsep}{3pt}
\caption{\textbf{Runtime efficiency metrics} on WebForge-Bench, averaged per task by difficulty level. ``Turns'' = LLM dialogue rounds; ``Acts'' = browser actions; ``Prompt'' and ``Compl'' = input/output tokens (in K). Models marked with $^\dagger$ do not support the step-level logging mode, resulting in systematically lower token counts; they are excluded from cross-model efficiency comparisons.}
\label{tab:runtime}
\resizebox{\textwidth}{!}{
\begin{tabular}{@{}lcccc!{\vrule width 0.8pt}cccc!{\vrule width 0.8pt}cccc@{}}
\toprule
\rowcolor{headercolor}
 & \multicolumn{4}{c!{\vrule width 0.8pt}}{\textbf{Level 1}} & \multicolumn{4}{c!{\vrule width 0.8pt}}{\textbf{Level 2}} & \multicolumn{4}{c}{\textbf{Level 3}} \\
\rowcolor{headercolor}
\textbf{Model} & \textbf{Turns} & \textbf{Acts} & \textbf{Prompt} & \textbf{Compl} & \textbf{Turns} & \textbf{Acts} & \textbf{Prompt} & \textbf{Compl} & \textbf{Turns} & \textbf{Acts} & \textbf{Prompt} & \textbf{Compl} \\
\midrule
\multicolumn{13}{@{}l}{\textit{(a) Multimodal (Screenshot + DOM)}} \\
\multicolumn{13}{@{}l}{\textit{\quad Closed-source}} \\
\rowcolor{geminicolor}
Gemini-3-Pro & 7.9 & 12.2 & 133K & 4.2K & 13.8 & 21.6 & 307K & 5.9K & 26.9 & 44.6 & 1036K & 11.2K \\
\rowcolor{geminicolor}
Gemini-3-Flash & 8.0 & 12.3 & 159K & 5.5K & 13.1 & 19.3 & 304K & 6.5K & 25.3 & 39.1 & 962K & 15.3K \\
\rowcolor{geminicolor}
Gemini-2.5-Flash-Lite$^\dagger$ & 12.0 & 6.6 & 224K & 4.6K & 16.5 & 11.5 & 254K & 3.4K & 26.1 & 21.9 & 520K & 5.6K \\
\rowcolor{claudecolor}
Claude-4.5-Sonnet & 11.0 & 12.3 & 260K & 3.8K & 18.7 & 20.7 & 591K & 6.9K & 33.8 & 37.4 & 1608K & 12.6K \\
\rowcolor{gptcolor}
GPT-5.2$^\dagger$ & 8.8 & 8.5 & 80K & 0.4K & 15.6 & 16.1 & 236K & 0.6K & 26.1 & 27.7 & 656K & 1.0K \\
\rowcolor{gptcolor}
GPT-5-Mini$^\dagger$ & 11.5 & 10.5 & 150K & 2.2K & 20.7 & 19.7 & 421K & 4.2K & 36.7 & 36.0 & 1164K & 9.7K \\
\rowcolor{gptcolor}
GPT-5-Nano$^\dagger$ & 18.1 & 13.7 & 277K & 9.4K & 29.3 & 23.3 & 590K & 19.5K & 38.4 & 30.8 & 892K & 31.3K \\
\multicolumn{13}{@{}l}{\textit{\quad Open-source}} \\
\rowcolor{kimicolor}
Kimi-K2.5 & 13.3 & 11.1 & 176K & 3.2K & 21.1 & 19.8 & 385K & 5.8K & 36.2 & 34.6 & 904K & 10.5K \\
\rowcolor{qwencolor}
Qwen3-VL-235B & 9.0 & 9.2 & 135K & 1.9K & 16.2 & 17.4 & 363K & 3.7K & 28.7 & 32.4 & 845K & 6.9K \\
\rowcolor{qwencolor}
Qwen3-Omni-30B$^\dagger$ & 34.3 & 6.9 & 463K & 4.4K & 43.2 & 6.8 & 641K & 6.6K & 46.8 & 8.0 & 740K & 7.1K \\
\midrule
\multicolumn{13}{@{}l}{\textit{(b) Text-only (DOM only)}} \\
\multicolumn{13}{@{}l}{\textit{\quad Text-only models}} \\
\rowcolor{deepseekcolor}
DeepSeek-V3.2 & 12.4 & 11.7 & 165K & 3.5K & 22.7 & 24.2 & 420K & 6.6K & 36.3 & 40.9 & 920K & 10.5K \\
\rowcolor{glmcolor}
GLM-4.7 & 11.6 & 12.8 & 138K & 3.7K & 22.7 & 25.6 & 376K & 7.5K & 34.4 & 40.2 & 761K & 11.5K \\
\multicolumn{13}{@{}l}{\textit{\quad Multimodal models (DOM only)}} \\
\rowcolor{geminicolor}
Gemini-3-Pro (T) & 10.6 & 16.8 & 144K & 5.4K & 21.6 & 33.9 & 412K & 8.9K & 33.7 & 57.7 & 875K & 13.2K \\
\rowcolor{geminicolor}
Gemini-3-Flash (T) & 10.5 & 15.4 & 213K & 7.5K & 29.8 & 47.1 & 854K & 26.1K & 41.4 & 65.5 & 1328K & 29.9K \\
\bottomrule
\end{tabular}}
\end{table*}

\cref{tab:runtime} reports the average runtime cost per task. Our evaluation framework requires the agent to explicitly output its observation, reasoning, and planned action at each step, providing an interpretable decision-making record; this logging incurs additional token consumption. The GPT series, Gemini-2.5-Flash-Lite, and Qwen3-Omni-30B do not adequately support this logging mode and skip step-level recording, reducing their token usage at the cost of trajectory interpretability. Because their token statistics are systematically lower for this structural reason, these models are marked with $^\dagger$ and excluded from the efficiency comparisons below.

\noindent\textbf{Cost scales super-linearly with difficulty.} From Level~1 to Level~3, prompt tokens grow 5--8$\times$ across all comparable models, reflecting longer trajectories and heavier context accumulation.

\noindent\textbf{Token efficiency varies across models.} Among comparable models, Gemini-3-Pro is the most prompt-efficient at Level~1 (133K), while Claude-4.5-Sonnet is the most expensive at Level~3 (1608K). At Level~2, Gemini-3-Flash (304K) and Gemini-3-Pro (307K) are comparably efficient, both well below Claude-4.5-Sonnet (591K).

\noindent\textbf{Qwen3-Omni-30B$^\dagger$ exhibits a unique failure mode:} it issues the fewest actions (6--8 per task) yet the most LLM turns (34--47), indicating frequent observation-only cycles without productive interactions.

\subsection{Per-Dimension Analysis}

\begin{table*}[t]
\centering
\caption{\textbf{Per-dimension accuracy (\%)} on WebForge-Bench. For each of the seven difficulty dimensions, we report accuracy at dimension level L1, L2, and L3. Text-only models and rows marked (T) receive DOM only (no screenshots).}
\label{tab:per_dimension}
\resizebox{\textwidth}{!}{%
\begin{tabular}{@{}l*{3}{c}!{\vrule width 0.8pt}*{3}{c}!{\vrule width 0.8pt}*{3}{c}!{\vrule width 0.8pt}*{3}{c}!{\vrule width 0.8pt}*{3}{c}!{\vrule width 0.8pt}*{3}{c}!{\vrule width 0.8pt}*{3}{c}@{}}
\toprule
\rowcolor{headercolor}
& \multicolumn{3}{c!{\vrule width 0.8pt}}{\textbf{Jump Depth}} & \multicolumn{3}{c!{\vrule width 0.8pt}}{\textbf{Jump Breadth}} & \multicolumn{3}{c!{\vrule width 0.8pt}}{\textbf{Page Interact.}} & \multicolumn{3}{c!{\vrule width 0.8pt}}{\textbf{Visual Compl.}} & \multicolumn{3}{c!{\vrule width 0.8pt}}{\textbf{Info Compl.}} & \multicolumn{3}{c!{\vrule width 0.8pt}}{\textbf{Reason./Calc}} & \multicolumn{3}{c}{\textbf{Risk Factor}} \\
\cmidrule(lr){2-4} \cmidrule(lr){5-7} \cmidrule(lr){8-10} \cmidrule(lr){11-13} \cmidrule(lr){14-16} \cmidrule(lr){17-19} \cmidrule(lr){20-22}
\rowcolor{headercolor}
\textbf{Model} & L1 & L2 & L3 & L1 & L2 & L3 & L1 & L2 & L3 & L1 & L2 & L3 & L1 & L2 & L3 & L1 & L2 & L3 & L1 & L2 & L3 \\
\midrule
\multicolumn{22}{@{}l}{\textit{(a) Multimodal (Screenshot + DOM)}} \\
\multicolumn{22}{@{}l}{\textit{\quad Closed-source}} \\
\rowcolor{geminicolor}
Gemini-3-Pro & \textbf{86.5} & \textbf{78.9} & \textbf{60.2} & \underline{84.8} & \textbf{79.9} & \textbf{51.2} & \textbf{84.0} & \textbf{74.9} & \textbf{65.0} & \textbf{90.8} & \textbf{78.9} & \textbf{55.8} & \textbf{84.7} & \textbf{75.7} & \textbf{53.2} & \textbf{91.4} & \textbf{74.6} & \textbf{58.3} & \textbf{80.6} & \textbf{70.3} & 23.1 \\
\rowcolor{geminicolor}
Gemini-3-Flash & 82.3 & 71.1 & 45.1 & 83.8 & 67.6 & 45.7 & 74.6 & 67.8 & 47.0 & 83.1 & 69.0 & 46.8 & \underline{81.2} & 64.0 & 39.0 & 84.7 & 68.3 & 42.6 & 72.2 & 60.0 & \textbf{38.5} \\
\rowcolor{geminicolor}
Gemini-2.5-Flash-Lite & 57.3 & 33.2 & 13.5 & 56.0 & 34.3 & 13.0 & 52.1 & 33.3 & 9.0 & 54.7 & 34.2 & 13.0 & 50.4 & 28.6 & 13.5 & 56.8 & 31.7 & 12.8 & 42.7 & 23.7 & 0.0 \\
\rowcolor{claudecolor}
Claude-4.5-Sonnet & \underline{85.8} & \underline{71.8} & \underline{50.0} & \textbf{85.9} & \underline{70.7} & \underline{48.1} & \underline{81.7} & \underline{69.2} & \underline{49.0} & 86.5 & 69.0 & \underline{51.5} & 81.2 & \underline{66.9} & \underline{48.9} & 87.4 & \underline{70.4} & \underline{46.8} & \underline{76.4} & \underline{60.9} & \underline{30.8} \\
\rowcolor{gptcolor}
GPT-5.2 & 79.2 & 62.9 & 33.5 & 76.4 & 62.8 & 27.8 & 71.8 & 58.1 & 42.0 & 84.5 & 58.1 & 31.9 & 74.0 & 58.1 & 25.5 & 86.0 & 59.0 & 26.4 & 67.3 & 48.6 & 15.4 \\
\rowcolor{gptcolor}
GPT-5-Mini & 81.2 & 66.1 & 29.7 & 82.2 & 63.0 & 25.3 & 80.8 & 59.4 & 23.0 & 83.7 & 62.7 & 31.2 & 77.2 & 56.4 & 27.7 & 84.7 & 61.8 & 26.8 & 71.1 & 44.3 & 23.1 \\
\rowcolor{gptcolor}
GPT-5-Nano & 61.8 & 26.1 & 5.6 & 59.2 & 28.7 & 7.4 & 61.5 & 25.4 & 3.0 & 50.1 & 27.8 & 12.6 & 47.2 & 24.3 & 9.9 & 51.2 & 30.9 & 6.4 & 40.3 & 17.7 & 0.0 \\
\multicolumn{22}{@{}l}{\textit{\quad Open-source}} \\
\rowcolor{kimicolor}
Kimi-K2.5 & 84.7 & 70.3 & 41.0 & 83.8 & 70.1 & 32.7 & 81.2 & 65.1 & 43.0 & 84.2 & \underline{71.5} & 40.9 & 79.9 & 62.6 & 41.8 & 86.4 & 67.3 & 39.1 & 75.0 & 54.3 & 15.4 \\
\rowcolor{qwencolor}
Qwen3-VL-235B & 72.2 & 48.9 & 21.4 & 70.7 & 49.1 & 19.1 & 69.0 & 46.1 & 18.0 & 73.9 & 44.7 & 21.9 & 63.0 & 45.0 & 19.1 & 75.1 & 45.5 & 18.7 & 58.7 & 32.3 & 23.1 \\
\rowcolor{qwencolor}
Qwen3-Omni-30B & 27.1 & 8.9 & 2.6 & 23.0 & 11.9 & 3.7 & 27.2 & 9.7 & 1.0 & 24.1 & 10.2 & 2.0 & 17.2 & 11.9 & 3.5 & 24.3 & 9.8 & 3.0 & 18.4 & 4.0 & 0.0 \\
\midrule
\multicolumn{22}{@{}l}{\textit{(b) Text-only (DOM only)}} \\
\multicolumn{22}{@{}l}{\textit{\quad Text-only models}} \\
\rowcolor{deepseekcolor}
DeepSeek-V3.2 & 76.4 & 45.8 & 23.3 & 71.7 & 48.9 & 21.6 & 58.2 & 51.2 & 14.0 & 81.7 & 39.8 & 19.3 & 67.3 & 42.4 & 19.1 & 79.4 & 43.0 & 19.6 & 56.2 & 38.0 & 15.4 \\
\rowcolor{glmcolor}
GLM-4.7 & 75.7 & 47.4 & 26.7 & 72.3 & 51.6 & 19.1 & 58.7 & 51.4 & 25.0 & 84.2 & 39.8 & 20.6 & 66.8 & 44.5 & 23.4 & 81.7 & 43.2 & 21.7 & 56.6 & 41.4 & 7.7 \\
\multicolumn{22}{@{}l}{\textit{\quad Multimodal models (DOM only)}} \\
\rowcolor{geminicolor}
Gemini-3-Pro (T) & 79.5 & 59.7 & 36.5 & 77.5 & 61.4 & 29.6 & 66.2 & 60.2 & 38.0 & \underline{87.4} & 56.7 & 28.9 & 74.0 & 55.2 & 31.9 & \underline{87.7} & 52.0 & 34.9 & 64.6 & 52.0 & 15.4 \\
\rowcolor{geminicolor}
Gemini-3-Flash (T) & 78.1 & 48.9 & 25.2 & 73.3 & 52.0 & 22.2 & 57.3 & 52.5 & 30.0 & 86.0 & 42.6 & 18.9 & 69.2 & 45.0 & 22.0 & 82.7 & 45.5 & 20.4 & 58.0 & 41.7 & 7.7 \\
\bottomrule
\end{tabular}%
}
\end{table*}

\cref{tab:per_dimension} provides a fine-grained breakdown of model accuracy across the seven difficulty dimensions at each dimension level. We also include text-only (T) variants for Gemini-3-Pro and Gemini-3-Flash.

\noindent\textbf{All dimensions exhibit monotonic difficulty scaling.} Across all models and dimensions, accuracy consistently decreases from L1 to L3, confirming that each dimension independently contributes to task difficulty. The steepest drops occur in \textit{Visual Complexity} (e.g., Gemini-3-Pro: 90.8\%$\rightarrow$55.8\%) and \textit{Jump Breadth} (84.8\%$\rightarrow$51.2\%), suggesting these dimensions are particularly discriminative.

\noindent\textbf{Reasoning/Calculation separates strong from weak models.} At L1, most models achieve $>$75\% on Reasoning/Calc, but at L3 a clear gap emerges: Gemini-3-Pro maintains 58.3\% while GPT-5-Nano drops to 6.4\%. This 52-point spread indicates that complex reasoning is a key differentiator for frontier models.

\noindent\textbf{Dimensions are correlated, yet discriminative.} The difficulty dimensions are not fully orthogonal: higher overall difficulty levels tend to elevate multiple dimensions simultaneously, attenuating inter-dimension performance differences. Nevertheless, the per-dimension breakdown still reveals meaningful capability distinctions---Visual Complexity produces the steepest L1-to-L3 drop while Reasoning/Calc best separates strong from weak models---confirming that multi-dimensional profiling provides diagnostic value beyond aggregate or per-level scores.

\subsection{Discussion}

\noindent\textbf{Comparison with existing benchmarks.}
\cref{tab:benchmark_comparison} positions WebForge-Bench among representative browser agent benchmarks along six key axes. Several observations emerge.
(1)~\emph{No prior benchmark simultaneously achieves realism, reproducibility, and scalability.} Real-website benchmarks (Mind2Web, WebVoyager, MMInA) offer realism but suffer from content drift---nearly half of the Mind2Web tasks expired within two years~\cite{xue2025onlinemind2web}, and WebVoyager reports 22.3\% of answers as non-deterministic. Controlled environments (WebArena, VisualWebArena, EntWorld) guarantee reproducibility but omit real-web noise and require costly manual curation. WebForge resolves this trilemma through automated generation of self-contained environments with injected real-web disturbances.
(2)~\emph{Difficulty control is absent or shallow.} WebArena, WebVoyager, and TheAgentCompany offer no difficulty stratification; VisualWebArena provides two post-hoc dimensions; WorkArena++ varies only instruction explicitness; EntWorld computes a post-hoc weighted score from SQL-structural features. WebForge is the first to offer seven a priori controllable dimensions, enabling fine-grained capability profiling (\cref{tab:per_dimension}).
(3)~\emph{Full automation remains rare.} Among all compared benchmarks, only EntWorld partially automates task generation (instantiation is automatic but environment deployment is manual); TheAgentCompany required ${\sim}$3{,}000 person-hours for 175 tasks. WebForge's four-agent pipeline generates complete interactive environments end-to-end with zero human annotation.
The accuracy range on WebForge-Bench (12.7--75.9\%) indicates that our benchmark is challenging yet solvable, avoiding the extremes of near-random scores or near-saturation.

\begin{table*}[t]
\centering
\scriptsize
\setlength{\tabcolsep}{3pt}
\caption{\textbf{Comparison with representative browser agent benchmarks.}
``Diff.\ Ctrl'' indicates whether task difficulty is controllable a priori during generation.
``Noise'' indicates systematic injection of real-web disturbances (pop-ups, cookie dialogs, network delays).
``Auto'' indicates fully automated environment and task generation with zero human annotation.
``Reprod.'' indicates whether the benchmark environment is self-contained and reproducible.}
\label{tab:benchmark_comparison}
\resizebox{\textwidth}{!}{%
\begin{tabular}{@{}llccccccc@{}}
\toprule
\rowcolor{headercolor}
\textbf{Benchmark} & \textbf{Type} & \textbf{\#Tasks} & \textbf{\#Dom.} & \textbf{Diff.\ Ctrl} & \textbf{Noise} & \textbf{Auto} & \textbf{Reprod.} & \textbf{Eval Paradigm} \\
\midrule
Mind2Web~\cite{deng2023mind2web} & Real & 2{,}350 & 137$^\dagger$ & \ding{55} & \ding{55}$^a$ & \ding{55} & \ding{55}$^b$ & Step-wise pred. \\
WebVoyager~\cite{he2024webvoyager} & Real & 643 & 15$^\dagger$ & \ding{55} & Passive & \ding{55} & \ding{55} & LMM-judge \\
MMInA~\cite{tian2025mmina} & Real & 1{,}050 & 14$^\dagger$ & \ding{55} & Passive & \ding{55} & \ding{55} & Hop SR \\
\midrule
WebArena~\cite{zhou2024webarena} & Ctrl & 812 & 4 & \ding{55} & \ding{55} & \ding{55} & \ding{51} & Programmatic \\
VisualWebArena~\cite{koh2024visualwebarena} & Ctrl & 910 & 3 & 2-dim$^c$ & \ding{55} & \ding{55} & \ding{51} & Hand-crafted \\
WorkArena++~\cite{drouin2024workarenaplusplus} & Ctrl & 682 & 1 & 1-dim$^d$ & \ding{55} & \ding{55} & \ding{51} & Oracle fn \\
EntWorld~\cite{mo2026entworld} & Ctrl & 1{,}756 & 6 & Post-hoc$^e$ & \ding{55} & Partial$^f$ & \ding{51} & SQL verification \\
TheAgentCompany~\cite{xu2025theagentcompany} & Ctrl & 175 & 1$^g$ & \ding{55} & Partial & \ding{55} & \ding{51} & Checkpoint \\
\midrule
\rowcolor{geminicolor}
\textbf{WebForge (Ours)} & \textbf{Auto} & \textbf{934} & \textbf{7} & \textbf{7-dim $\times$ 3} & \ding{51} & \ding{51} & \ding{51} & \textbf{Final-state} \\
\bottomrule
\end{tabular}}
{\raggedright\footnotesize
$^\dagger$Number of websites, not thematic domains.
$^a$Annotation protocol explicitly excludes pop-ups and CAPTCHAs.
$^b$${\sim}$50\% of tasks expired within two years~\cite{xue2025onlinemind2web}.
$^c$Action difficulty + visual difficulty, annotated post hoc.
$^d$Controls instruction explicitness only (explicit steps vs.\ ticket+KB).
$^e$$D_{\text{task}}=\sum_i w_i d_i$ over 5 SQL-structural dimensions, computed post hoc.
$^f$Task instantiation automated; environment deployment and data population manual.
$^g$Single simulated company with 7 job categories.\par}
\end{table*}

\noindent\textbf{Limitations.} This work focuses on using the WebForge pipeline for automated benchmark generation. However, high-quality training data is equally critical for advancing browser agents, and large-scale collection of such data remains a core bottleneck in the field. WebForge's automated environment construction and task generation capabilities are naturally suited to training data production---a promising future direction is to extend the pipeline into a framework for generating high-quality browser agent training data.

\section{Conclusion}

We presented WebForge, the first fully automated framework for constructing realistic, reproducible, and scalable browser agent benchmarks. By decomposing benchmark generation into a four-agent pipeline (Plan $\rightarrow$ Generate $\rightarrow$ Refine $\rightarrow$ Validate) with a seven-dimensional difficulty control framework, WebForge resolves the benchmark trilemma that constrains existing approaches. Experiments on WebForge-Bench (934 tasks, 7 domains $\times$ 3 levels) reveal that difficulty-level stratification effectively separates model capabilities, cross-domain analysis uncovers capability biases invisible to aggregate scores, and visual input contributes 14--16 points of accuracy. Pipeline ablations confirm that each stage---plan refinement and rule-driven quality assurance---is essential for producing high-quality tasks. We believe WebForge provides a scalable foundation for continuously evolving browser agent evaluation as agent capabilities advance.

\bibliographystyle{splncs04}
\bibliography{main}

@String(CVPR  = {IEEE Conf. Comput. Vis. Pattern Recog.})

@String(NeurIPS = {Adv. Neural Inform. Process. Syst.})

@String(ICML  = {Int. Conf. Mach. Learn.})

@String(ICLR  = {Int. Conf. Learn. Represent.})

@String(CVPR  = {CVPR})

@String(NeurIPS = {NeurIPS})

@String(ICML  = {ICML})

@String(ICLR  = {ICLR})

@inproceedings{zhou2024webarena,
  title={{WebArena}: A Realistic Web Environment for Building Autonomous Agents},
  author={Zhou, Shuyan and Xu, Frank F. and Zhu, Hao and Zhou, Xuhui and Lo, Robert and Sridhar, Abishek and Cheng, Xianyi and Ou, Tianyue and Bisk, Yonatan and Fried, Daniel and Alon, Uri and Neubig, Graham},
  booktitle=ICLR,
  year={2024}
}

@inproceedings{he2024webvoyager,
  title={{WebVoyager}: Building an End-to-End Web Agent with Large Multimodal Models},
  author={He, Hongliang and Yao, Wenlin and Ma, Kaixin and Yu, Wenhao and Dai, Yong and Zhang, Hongming and Lan, Zhenzhong and Yu, Dong},
  booktitle={ACL},
  year={2024}
}

@inproceedings{deng2023mind2web,
  title={{Mind2Web}: Towards a Generalist Agent for the Web},
  author={Deng, Xiang and Gu, Yu and Zheng, Boyuan and Chen, Shijie and Stevens, Samuel and Wang, Boshi and Sun, Huan and Su, Yu},
  booktitle={NeurIPS Datasets and Benchmarks Track},
  year={2023}
}

@article{wang2026colorbrowseragent,
  title={{ColorBrowserAgent}: Complex Long-Horizon Browser Agent with Adaptive Knowledge Evolution},
  author={Wang, Jihong and Zhou, Jiamu and Zhang, Weiming and Liu, Weiwen and Zhang, Zhuosheng and Lou, Xingyu and Zhang, Weinan and Deng, Huarong and Wang, Jun},
  journal={arXiv preprint arXiv:2601.07262},
  year={2026}
}

@inproceedings{wei2025webagentr1,
  title={{WebAgent-R1}: Training Web Agents via End-to-End Multi-Turn Reinforcement Learning},
  author={Wei, Zhepei and Yao, Wenlin and Liu, Yao and Zhang, Weizhi and Lu, Qin and Qiu, Liang and Yu, Changlong and Xu, Puyang and Zhang, Chao and Yin, Bing and Yun, Hyokun and Li, Lihong},
  booktitle={EMNLP},
  year={2025}
}

@inproceedings{xue2025onlinemind2web,
  title={An Illusion of Progress? Assessing the Current State of Web Agents},
  author={Xue, Tianci and Qi, Weijian and Shi, Tianneng and Song, Chan Hee and Gou, Boyu and Song, Dawn and Sun, Huan and Su, Yu},
  booktitle={COLM},
  year={2025}
}

@inproceedings{drouin2024workarena,
  title={{WorkArena}: How Capable Are Web Agents at Solving Common Knowledge Work Tasks?},
  author={Drouin, Alexandre and Gasse, Maxime and Caccia, Massimo and Laradji, Issam H. and Del Verme, Manuel and Marty, Tom and Boisvert, L{\'e}o and Thakkar, Megh and Cappart, Quentin and Vazquez, David and Chapados, Nicolas and Lacoste, Alexandre},
  booktitle=ICML,
  year={2024}
}

@article{mo2026entworld,
  title={{EntWorld}: A Holistic Environment and Benchmark for Verifiable Enterprise {GUI} Agents},
  author={Mo, Ying and Bai, Yu and Sun, Dapeng and Shi, Yuqian and Miao, Yukai and Chen, Li and Li, Dan},
  journal={arXiv preprint arXiv:2601.17722},
  year={2026}
}

@inproceedings{butt2025benchagents,
  title={{BenchAgents}: Automated Benchmark Creation with Agent Interaction},
  author={Butt, Natasha and Chandrasekaran, Varun and Joshi, Neel and Nushi, Besmira and Balachandran, Vidhisha},
  booktitle={ICLR 2025 Workshop on Navigating and Addressing Data Problems for Foundation Models (DATA-FM)},
  year={2025}
}

@inproceedings{li2025autobencher,
  title={{AutoBencher}: Towards Declarative Benchmark Construction},
  author={Li, Xiang Lisa and Kaiyom, Farzaan and Liu, Evan Zheran and Mai, Yifan and Liang, Percy and Hashimoto, Tatsunori},
  booktitle=ICLR,
  year={2025}
}

@inproceedings{zhu2024dyval,
  title={{DyVal}: Dynamic Evaluation of Large Language Models for Reasoning Tasks},
  author={Zhu, Kaijie and Chen, Jiaao and Wang, Jindong and Gong, Neil Zhenqiang and Yang, Diyi and Xie, Xing},
  booktitle=ICLR,
  year={2024}
}

@inproceedings{sun2025osgenesis,
  title={{OS-Genesis}: Automating {GUI} Agent Trajectory Construction via Reverse Task Synthesis},
  author={Sun, Qiushi and Cheng, Kanzhi and Ding, Zichen and Jin, Chuanyang and Wang, Yian and Xu, Fangzhi and Wu, Zhenyu and Jia, Chengyou and Chen, Liheng and Liu, Zhoumianze and Kao, Ben and Li, Guohao and He, Junxian and Qiao, Yu and Wu, Zhiyong},
  booktitle={ACL},
  year={2025}
}

@inproceedings{liu2024visualwebbench,
  title={{VisualWebBench}: How Far Have Multimodal {LLMs} Evolved in Web Page Understanding and Grounding?},
  author={Liu, Junpeng and Song, Yifan and Lin, Bill Yuchen and Lam, Wai and Neubig, Graham and Li, Yuanzhi and Yue, Xiang},
  booktitle={COLM},
  year={2024}
}

@inproceedings{shlomov2025grounding2planning,
  title={From Grounding to Planning: Benchmarking Bottlenecks in Web Agents},
  author={Shlomov, Segev and Wiesel, Ben and Sela, Aviad and Levy, Ido and Galanti, Liane and Abitbol, Roy},
  booktitle={ECAI},
  year={2025}
}

@inproceedings{koh2024visualwebarena,
  title={{VisualWebArena}: Evaluating Multimodal Agents on Realistic Visual Web Tasks},
  author={Koh, Jing Yu and Lo, Robert and Jang, Lawrence and Duvvur, Vikram and Lim, Ming and Huang, Po-Yu and Neubig, Graham and Zhou, Shuyan and Salakhutdinov, Russ and Fried, Daniel},
  booktitle={ACL},
  year={2024}
}

@inproceedings{drouin2024workarenaplusplus,
  title={{WorkArena++}: Towards Compositional Planning and Reasoning-based Common Knowledge Work Tasks},
  author={Boisvert, L\'{e}o and Thakkar, Megh and Gasse, Maxime and Caccia, Massimo and Le Sellier De Chezelles, Thibault and Cappart, Quentin and Chapados, Nicolas and Lacoste, Alexandre and Drouin, Alexandre},
  booktitle=NeurIPS,
  year={2024}
}

@inproceedings{levy2026stwebagentbench,
  title={{ST-WebAgentBench}: A Benchmark for Evaluating Safety and Trustworthiness in Web Agents},
  author={Levy, Ido and Wiesel, Ben and Marreed, Sami and Oved, Alon and Yaeli, Avi and Shlomov, Segev},
  booktitle=ICLR,
  year={2026}
}

@article{pan2024webcanvas,
  title={{WebCanvas}: Benchmarking Web Agents in Online Environments},
  author={Pan, Yichen and Kong, Dehan and Zhou, Sida and Cui, Cheng and Leng, Yifei and Jiang, Bing and Liu, Hangyu and Shang, Yanyi and Zhou, Shuyan and Wu, Tongshuang and Wu, Zhengyang},
  journal={arXiv preprint arXiv:2406.12373},
  year={2024}
}

@inproceedings{zhu2024dyval2,
  title={Dynamic Evaluation of Large Language Models by Meta Probing Agents},
  author={Zhu, Kaijie and Wang, Jindong and Zhao, Qinlin and Xu, Ruochen and Xie, Xing},
  booktitle=ICML,
  year={2024}
}

@inproceedings{shen2024taskbench,
  title={{TaskBench}: Benchmarking Large Language Models for Task Automation},
  author={Shen, Yongliang and Song, Kaitao and Tan, Xu and Zhang, Wenqi and Ren, Kan and Yuan, Siyu and Lu, Weiming and Li, Dongsheng and Zhuang, Yueting},
  booktitle=NeurIPS,
  year={2024}
}

@inproceedings{hong2024cogagent,
  title={{CogAgent}: A Visual Language Model for {GUI} Agents},
  author={Hong, Wenyi and Wang, Weihan and Lv, Qingsong and Xu, Jiazheng and Yu, Wenmeng and Ji, Junhui and Wang, Yan and Wang, Zihan and Zhang, Yuxuan and Li, Juanzi and Xu, Bin and Dong, Yuxiao and Ding, Ming and Tang, Jie},
  booktitle=CVPR,
  year={2024}
}

@article{wei2025browsecomp,
  title={{BrowseComp}: A Simple Yet Challenging Benchmark for Browsing Agents},
  author={Wei, Jason and Sun, Zhiqing and Papay, Spencer and McKinney, Scott and Han, Jeffrey and Fulford, Isa and Chung, Hyung Won and Passos, Alex Tachard and Fedus, William and Glaese, Amelia},
  journal={arXiv preprint arXiv:2504.12516},
  year={2025}
}

@inproceedings{mialon2024gaia,
  title={{GAIA}: A Benchmark for General {AI} Assistants},
  author={Mialon, Gr{\'e}goire and Fourrier, Cl{\'e}mentine and Wolf, Thomas and LeCun, Yann and Scialom, Thomas},
  booktitle=ICLR,
  year={2024}
}

@article{anupam2025browserarena,
  title={{BrowserArena}: Evaluating {LLM} Agents on Real-World Web Navigation Tasks},
  author={Anupam, Sagnik and Brown, Davis and Li, Shuo and Wong, Eric and Hassani, Hamed and Bastani, Osbert},
  journal={arXiv preprint arXiv:2510.02418},
  year={2025}
}

@article{miyai2025webchorearena,
  title={{WebChoreArena}: Evaluating Web Browsing Agents on Realistic Tedious Web Tasks},
  author={Miyai, Atsuyuki and Zhao, Zaiying and Egashira, Kazuki and Sato, Atsuki and Sunada, Tatsumi and Onohara, Shota and Yamanishi, Hiromasa and Toyooka, Mashiro and Nishina, Kunato and Maeda, Ryoma and Aizawa, Kiyoharu and Yamasaki, Toshihiko},
  journal={arXiv preprint arXiv:2506.01952},
  year={2025}
}

@inproceedings{tian2025mmina,
  title={{MMInA}: Benchmarking Multihop Multimodal Internet Agents},
  author={Tian, Shulin and Zhang, Ziniu and Chen, Liangyu and Liu, Ziwei},
  booktitle={Findings of ACL},
  year={2025}
}

@inproceedings{xu2025theagentcompany,
  title={{TheAgentCompany}: Benchmarking {LLM} Agents on Consequential Real World Tasks},
  author={Xu, Frank F. and Song, Yufan and Li, Boxuan and Tang, Yuxuan and Jain, Kritanjali and Bao, Mengxue and Wang, Zora Zhiruo and Zhou, Xuhui and Guo, Zhitong and Cao, Murong and Yang, Mingyang and Lu, Hao Yang and Martin, Amaad and Su, Zhe and Maben, Leander Melroy and Mehta, Raj and Chi, Wayne and Jang, Lawrence Keunho and Xie, Yiqing and Zhou, Shuyan and Neubig, Graham},
  booktitle={NeurIPS Datasets and Benchmarks Track},
  year={2025}
}

@misc{gemini3pro2025,
  title = {A New Era of Intelligence with {Gemini 3}},
  author = {{Google DeepMind}},
  howpublished = {\url{https://blog.google/products-and-platforms/products/gemini/gemini-3}},
  year = {2025},
  note = {Accessed: 2026-03-04}
}

@misc{gemini3flash2025,
  title = {{Gemini 3 Flash}: Frontier Intelligence Built for Speed},
  author = {{Google DeepMind}},
  howpublished = {\url{https://blog.google/products-and-platforms/products/gemini/gemini-3-flash}},
  year = {2025},
  note = {Accessed: 2026-03-04}
}

@misc{google2025gemini25flashlite,
  title = {We're Expanding Our {Gemini 2.5} Family of Models},
  author = {{Google DeepMind}},
  howpublished = {\url{https://blog.google/products-and-platforms/products/gemini/gemini-2-5-model-family-expands}},
  year = {2025},
  note = {Accessed: 2026-03-04}
}

@misc{anthropic2025claude45sonnet,
  title = {Introducing {Claude Sonnet 4.5}},
  author = {{Anthropic}},
  howpublished = {\url{https://anthropic.com/news/claude-sonnet-4-5}},
  year = {2025},
  note = {Accessed: 2026-03-04}
}

@misc{openai2025gpt52,
  title = {{GPT-5.2}: The Best Model for Coding and Agentic Tasks},
  author = {{OpenAI}},
  howpublished = {\url{https://developers.openai.com/api/docs/models/gpt-5.2}},
  year = {2025},
  note = {Accessed: 2026-03-04}
}

@misc{openai2025gpt5mini,
  title = {{GPT-5 Mini}: A Faster, Cost-Efficient Version of {GPT-5}},
  author = {{OpenAI}},
  howpublished = {\url{https://developers.openai.com/api/docs/models/gpt-5-mini}},
  year = {2025},
  note = {Accessed: 2026-03-04}
}

@misc{openai2025gpt5nano,
  title = {{GPT-5 Nano}: Fastest, Most Cost-Efficient Version of {GPT-5}},
  author = {{OpenAI}},
  howpublished = {\url{https://developers.openai.com/api/docs/models/gpt-5-nano}},
  year = {2025},
  note = {Accessed: 2026-03-04}
}

@misc{zhipuai2025glm47,
  title = {{GLM-4.7}: Comprehensive Coding Capability Enhancement},
  author = {{Zhipu AI}},
  howpublished = {\url{https://docs.z.ai/guides/llm/glm-4.7}},
  year = {2025},
  note = {Accessed: 2026-03-04}
}

@misc{deepseekai2025deepseekv32,
  title = {{DeepSeek-V3.2}: Reasoning-First Models Built for Agents},
  author = {{DeepSeek-AI}},
  howpublished = {\url{https://api-docs.deepseek.com/news/news251201}},
  year = {2025},
  note = {Accessed: 2026-03-04}
}

@misc{moonshotai2026kimik25,
  title = {{Kimi K2.5}: Open-Source Native Multimodal Agentic Model},
  author = {{Moonshot AI}},
  howpublished = {\url{https://github.com/MoonshotAI/Kimi-K2.5}},
  year = {2026},
  note = {Accessed: 2026-03-04}
}

@misc{qwen2025qwen3vl,
  title = {{Qwen3-VL}: The Most Powerful Vision-Language Model in the Qwen Series},
  author = {{Qwen Team, Alibaba Cloud}},
  howpublished = {\url{https://github.com/QwenLM/Qwen3-VL}},
  year = {2025},
  note = {Accessed: 2026-03-04}
}

@misc{qwen2025qwen3omni,
  title = {{Qwen3-Omni}: Natively Omni-Modal Foundation Models},
  author = {{Qwen Team, Alibaba Cloud}},
  howpublished = {\url{https://github.com/QwenLM/Qwen3-Omni}},
  year = {2025},
  note = {Accessed: 2026-03-04}
}

@inproceedings{zheng2024seeact,
  title={{GPT-4V(ision)} is a Generalist Web Agent, if Grounded},
  author={Zheng, Boyuan and Gou, Boyu and Kil, Jihyung and Sun, Huan and Su, Yu},
  booktitle=ICML,
  year={2024}
}

@inproceedings{yang2025agentoccam,
  title={{AgentOccam}: A Simple Yet Strong Baseline for {LLM}-Based Web Agents},
  author={Yang, Ke and Liu, Yao and Chaudhary, Sapana and Fakoor, Rasool and Chaudhari, Pratik and Karypis, George and Rangwala, Huzefa},
  booktitle=ICLR,
  year={2025}
}

@inproceedings{yao2023react,
  title={{ReAct}: Synergizing Reasoning and Acting in Language Models},
  author={Yao, Shunyu and Zhao, Jeffrey and Yu, Dian and Du, Nan and Shafran, Izhak and Narasimhan, Karthik and Cao, Yuan},
  booktitle=ICLR,
  year={2023}
}

\clearpage

% =====================================================
% Supplementary Material
% =====================================================
\newpage

\begin{center}
{\LARGE\bfseries Supplementary Material}\\[6pt]
{\large WebForge: Breaking the Realism-Reproducibility-Scalability Trilemma in Browser Agent Benchmark}
\end{center}
\vspace{12pt}

% Reset section counter to use letters
\setcounter{section}{0}
\renewcommand{\thesection}{\Alph{section}}

\section{Statistical Validation of Difficulty Dimensions}
\label{app:stats}

We conduct statistical analyses on the 934-task annotations to validate the seven-dimensional framework. We first define the seven difficulty dimensions and their level criteria, then present distribution, correlation, and solvability analyses.

\subsection{Difficulty Dimension Definitions}
\label{app:dim_def}

WebForge defines seven orthogonal dimensions to characterize the difficulty of web-based tasks. Each dimension is graded on three levels (L1--L3) with increasing complexity. Table~\ref{tab:dim_def} provides the complete definitions.

\begin{table*}[!b]
\centering
\scriptsize
\caption{\textbf{Seven difficulty dimensions and their level definitions.} Each dimension captures a distinct axis of task complexity; L1 represents the easiest and L3 the hardest level.}
\label{tab:dim_def}
\begin{tabularx}{\linewidth}{@{}p{2.4cm}|X|X|X@{}}
\toprule
\rowcolor{headercolor}
\textbf{Dimension} & \textbf{L1 (Easy)} & \textbf{L2 (Medium)} & \textbf{L3 (Hard)} \\
\midrule
\textbf{Jump Depth} \newline {\scriptsize(Page transitions from start to finish)} & 1--2 page transitions & 3--5 page transitions & 6+ page transitions \\
\midrule
\textbf{Jump Breadth} \newline {\scriptsize(Max options to evaluate on a single page)} & 1--2 links/options & 3--5 links/options & 6+ links/options \\
\midrule
\textbf{Page Interaction} \newline {\scriptsize(Actions per page: forms, popups, buttons)} & 1--2 interactions (simple clicks, single field) & 3--5 interactions (multi-field forms, dropdowns) & 6+ interactions (complex forms, multi-step wizards) \\
\midrule
\textbf{Visual Complexity} \newline {\scriptsize(Reliance on visual reasoning)} & All information in text/DOM; no visual parsing needed & Simple chart/image reading; single visual element & Complex visual reasoning; correlating multiple charts/images \\
\midrule
\textbf{Info Complexity} \newline {\scriptsize(Volume of information to process)} & Key information is prominent and easy to find & Moderate density; requires scanning & High density; long documents; information buried in noise \\
\midrule
\textbf{Reasoning/Calc} \newline {\scriptsize(Reasoning and calculation complexity)} & Direct lookup; no calculation needed & Simple arithmetic; basic comparison/filtering & Multi-step reasoning; optimization; complex math \\
\midrule
\textbf{Risk Factor} \newline {\scriptsize(Irreversible operation risk)} & Read-only operations; no risk & Irreversible but with clear confirmation dialogs & Subtle irreversible actions; no obvious warnings \\
\bottomrule
\end{tabularx}
\end{table*}

\noindent\textbf{Overall difficulty levels.} The seven per-dimension levels are aggregated into three overall difficulty levels via strict compositional rules:
\begin{itemize}[nosep]
\item \textbf{Level~1}: At most 2 dimensions at L2; all others at L1; no L3 allowed.
\item \textbf{Level~2}: At least 2 dimensions at L2; at most 1 dimension at L3.
\item \textbf{Level~3}: At least 2 dimensions at L3 \emph{and} at least 2 at L2.
\end{itemize}

\noindent These rules ensure that higher overall levels demand multi-faceted complexity rather than a single extreme dimension, producing tasks that stress diverse agent capabilities simultaneously.

\subsection{Dimension Distribution}
\label{app:dim_dist}

\cref{tab:dim_dist} shows the distribution of tasks across dimension levels. Most dimensions have a roughly balanced distribution, though some exhibit natural skew. Risk Factor is heavily concentrated at L1 (61.1\%) with only 13 tasks (1.4\%) at L3: since L3 requires irreversible operations with no obvious warnings (e.g., permanent account deletion without confirmation dialogs), such scenarios are inherently rare in real-world web applications, so the Plan Agent naturally generates far more low-risk (L1) or moderately risky (L2, with clear warnings) tasks. Page Interaction concentrates at L2 (66.5\%) because most web tasks naturally involve moderate form/button interactions. We note that the WebForge framework supports custom dimension-level constraints (e.g., forcing Risk Factor to L3 for safety-focused evaluation), enabling users to generate domain-specific benchmarks; the distribution reported here reflects the \emph{general-purpose} setting without such overrides, which better represents the natural distribution of real-world web tasks.

\begin{table}[tbp]
\centering
\scriptsize
\caption{\textbf{Dimension-level distribution} across 934 tasks. Each cell shows count (\%).}
\label{tab:dim_dist}
\begin{tabular}{@{}lccc@{}}
\toprule
\rowcolor{headercolor}
\textbf{Dimension} & \textbf{L1} & \textbf{L2} & \textbf{L3} \\
\midrule
Jump Depth & 288 (30.8\%) & 380 (40.7\%) & 266 (28.5\%) \\
Jump Breadth & 191 (20.4\%) & 581 (62.2\%) & 162 (17.3\%) \\
Page Interaction & 213 (22.8\%) & 621 (66.5\%) & 100 (10.7\%) \\
Visual Complexity & 349 (37.4\%) & 284 (30.4\%) & 301 (32.2\%) \\
Info Complexity & 373 (39.9\%) & 420 (45.0\%) & 141 (15.1\%) \\
Reasoning/Calc & 301 (32.2\%) & 398 (42.6\%) & 235 (25.2\%) \\
Risk Factor & 571 (61.1\%) & 350 (37.5\%) & 13 (1.4\%) \\
\bottomrule
\end{tabular}
\end{table}

\subsection{Domain $\times$ Level Distribution and Pipeline Pass Rate}
\label{app:domain_dist}

The WebForge pipeline generates 60 candidate tasks per domain--level pair (7 domains $\times$ 3 levels = 1{,}260 total). After Validation Agent filtering, 934 tasks pass solvability verification. \cref{tab:domain_dist} shows the number of passing tasks and the corresponding pass rate for each cell.

\begin{table}[tbp]
\centering
\scriptsize
\caption{\textbf{Task count and pipeline pass rate (\%)} by domain and difficulty level. Each cell shows \emph{count} (\emph{rate}), where rate = count/60.}
\label{tab:domain_dist}
\begin{tabular}{@{}lcccc@{}}
\toprule
\rowcolor{headercolor}
\textbf{Domain} & \textbf{L1} & \textbf{L2} & \textbf{L3} & \textbf{Total} \\
\midrule
D1: Consumer Trans. & 39 (65.0) & 41 (68.3) & 35 (58.3) & 115 (63.9) \\
D2: Content Moder. & 39 (65.0) & 48 (80.0) & 38 (63.3) & 125 (69.4) \\
D3: Enterprise Proc. & 43 (71.7) & 42 (70.0) & 46 (76.7) & 131 (72.8) \\
D4: Info Retrieval & 53 (88.3) & 58 (96.7) & 49 (81.7) & 160 (88.9) \\
D5: Platform Mgmt. & 41 (68.3) & 50 (83.3) & 40 (66.7) & 131 (72.8) \\
D6: Tool Usage & 42 (70.0) & 51 (85.0) & 48 (80.0) & 141 (78.3) \\
D7: Content Creation & 44 (73.3) & 50 (83.3) & 37 (61.7) & 131 (72.8) \\
\midrule
\textbf{Total} & 301 (71.7) & 340 (81.0) & 293 (69.8) & 934 (74.1) \\
\bottomrule
\end{tabular}
\end{table}

\noindent The overall pipeline pass rate is \textbf{74.1\%} (934/1{,}260). By difficulty level, Level~2 achieves the highest pass rate (81.0\%), followed by Level~1 (71.7\%) and Level~3 (69.8\%). By domain, Info Retrieval (D4) has the highest pass rate (88.9\%) and Consumer Transaction (D1) the lowest (63.9\%).

\subsection{Inter-Dimension Correlation Analysis}
\label{app:correlation}

\begin{figure}[tbp]
\centering
\includegraphics[width=0.85\linewidth]{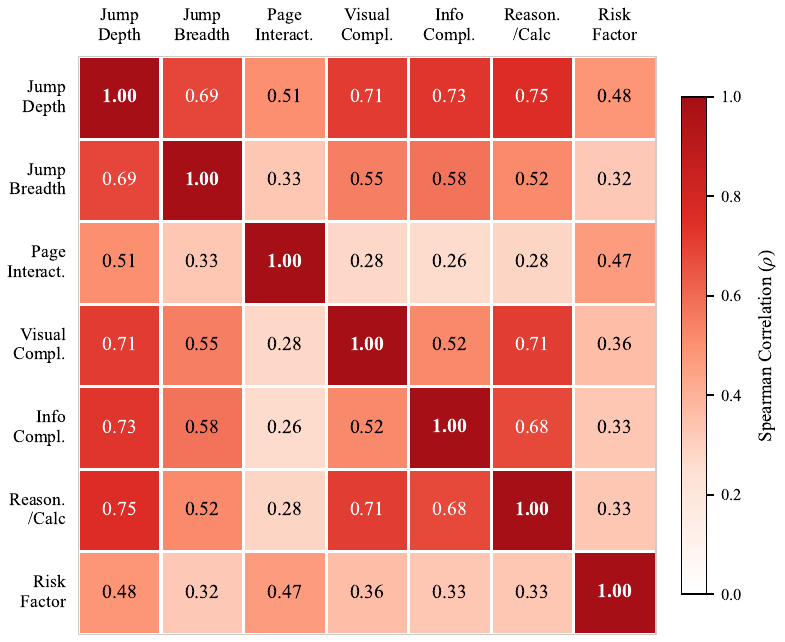}
\caption{\textbf{Spearman rank correlation matrix} ($\rho$) between the seven difficulty dimensions. The average off-diagonal $|\rho| = 0.495$, indicating moderate positive correlation driven by the overall difficulty constraints, while preserving sufficient independence for discriminative profiling.}
\label{fig:corr}
\end{figure}

To assess whether the seven dimensions provide independent capability signals, we compute pairwise Spearman rank correlations across all 934 tasks (\cref{fig:corr}).

\noindent\textbf{Key findings.} (1)~The average off-diagonal $|\rho| = 0.495$ (range: 0.262--0.754), indicating moderate positive correlation. This is expected because the overall difficulty constraints (e.g., Level~3 requires $\geq$2 dimensions at L3) induce co-variation. (2)~The strongest correlation is between Jump Depth and Reasoning/Calc ($\rho = 0.754$), reflecting that deeper navigation tasks naturally require more complex reasoning. (3)~Page Interaction shows the weakest correlations with other dimensions (avg.\ $|\rho| = 0.354$), suggesting it captures a relatively independent capability axis---interaction complexity (forms, popups, buttons) varies largely independently of navigation depth or visual complexity. (4)~Risk Factor also shows low correlations (avg.\ $|\rho| = 0.383$), consistent with irreversible operations being a distinct challenge orthogonal to other dimensions.

Despite the moderate inter-dimension correlations, the per-dimension accuracy analysis in the main paper (Table~4) demonstrates that different dimensions produce distinct capability profiles---for example, Visual Complexity produces the steepest L1-to-L3 accuracy drop while Reasoning/Calc best separates strong from weak models---confirming diagnostic value beyond what aggregate scoring provides.

\subsection{Task Solvability Analysis}
\label{app:solvability}

We analyze the solvability of WebForge-Bench tasks: a task is considered ``solved'' if \emph{at least one} of the 14 evaluated models (12 distinct models plus 2 text-only variants) answers it correctly. This metric reflects the intrinsic difficulty ceiling of the benchmark---tasks that no model can solve represent the hardest frontier.

\noindent\textbf{Results by difficulty level.}
\begin{itemize}
\item \textbf{Level~1}: 286/301 solved (\textbf{95.0\%}), with an average of 10.4 models solving each task.
\item \textbf{Level~2}: 317/340 solved (\textbf{93.2\%}), with an average of 7.9 models per task.
\item \textbf{Level~3}: 224/293 solved (\textbf{76.5\%}), with an average of only 4.3 models per task.
\item \textbf{Overall}: 827/934 solved (\textbf{88.5\%}).
\end{itemize}

\noindent\textbf{Analysis.} The solvability rate decreases monotonically from Level~1 to Level~3, confirming that the difficulty control framework produces progressively harder tasks. At Level~1, nearly all tasks (95.0\%) are solvable and most models succeed (avg.\ 10.4/14), indicating appropriate accessibility. At Level~3, roughly one-quarter of tasks (23.5\%) remain unsolved by all 14 models, and the average number of successful models drops sharply to 4.3, demonstrating strong discriminative power.

\noindent\textbf{Unsolved task distribution.} Among the 107 unsolved tasks, 69 (64.5\%) are Level~3, 23 (21.5\%) are Level~2, and 15 (14.0\%) are Level~1. By domain, Content Moderation (D2, 84.8\% solvability) and Platform Management (D5, 84.7\%) have the lowest solvability rates, while Content Creation (D7, 92.4\%) and Consumer Transaction (D1, 91.3\%) are most solvable---consistent with the cross-domain difficulty patterns identified in the main paper.

\section{Dimension-Level Accuracy Analysis}
\label{app:accuracy}

We provide detailed per-dimension accuracy drop analysis for representative models, complementing the summary in the main paper (Table~4).

\cref{tab:dim_drop} shows the L1-to-L3 accuracy drop ($\Delta$) for three models spanning the performance spectrum: Gemini-3-Pro (strongest, 75.9\%), Gemini-3-Flash (mid-tier, 67.1\%), and GPT-5-Nano (weakest, 31.3\%).

\begin{table*}[tbp]
\centering
\scriptsize
\setlength{\tabcolsep}{3pt}
\caption{\textbf{Per-dimension accuracy drop} (L1 $\rightarrow$ L3, in pp) for three representative models. Larger $\Delta$ = more discriminative.}
\label{tab:dim_drop}
\begin{tabular}{@{}lcrrrrrrr@{}}
\toprule
\rowcolor{headercolor}
\textbf{Model} & & \textbf{J.Dep.} & \textbf{J.Bre.} & \textbf{Pg.Int.} & \textbf{Vis.C.} & \textbf{Inf.C.} & \textbf{Rea.C.} & \textbf{Risk} \\
\midrule
\multirow{3}{*}{Gemini-3-Pro} 
 & L1 & 86.5 & 84.8 & 84.0 & 90.8 & 84.7 & 91.4 & 80.6 \\
 & L3 & 60.2 & 51.2 & 65.0 & 55.8 & 53.2 & 58.3 & 23.1 \\
 & $\Delta$ & 26.3 & 33.6 & 19.0 & 35.0 & 31.5 & 33.1 & 57.5 \\
\midrule
\multirow{3}{*}{Gemini-3-Flash}
 & L1 & 82.3 & 83.8 & 74.6 & 83.1 & 81.2 & 84.7 & 72.2 \\
 & L3 & 45.1 & 45.7 & 47.0 & 46.8 & 39.0 & 42.6 & 38.5 \\
 & $\Delta$ & 37.2 & 38.1 & 27.6 & 36.3 & 42.2 & 42.2 & 33.7 \\
\midrule
\multirow{3}{*}{GPT-5-Nano}
 & L1 & 61.8 & 59.2 & 61.5 & 50.1 & 47.2 & 51.2 & 40.3 \\
 & L3 & 5.6 & 7.4 & 3.0 & 12.6 & 9.9 & 6.4 & 0.0 \\
 & $\Delta$ & 56.2 & 51.8 & 58.5 & 37.5 & 37.3 & 44.8 & 40.3 \\
\bottomrule
\end{tabular}
\end{table*}

\noindent\textbf{Key observations.} (1)~\textit{Risk Factor shows the largest absolute drop for strong models}: Gemini-3-Pro drops 57.5pp on Risk Factor (80.6\%$\rightarrow$23.1\%), but this is partly driven by the extremely small L3 sample size (only 13 tasks). (2)~\textit{Weak models show near-zero L3 accuracy across all dimensions}: GPT-5-Nano achieves 0--12.6\% on all dimensions at L3, indicating a general capability floor rather than dimension-specific weaknesses; the variation in $\Delta$ for this model primarily reflects differences in L1 starting points. (3)~\textit{Info Complexity and Reasoning/Calc show the largest drops for mid-tier models}: Gemini-3-Flash drops 42.2pp on both, indicating that information processing capacity and reasoning ability are key differentiators in the mid-performance range. (4)~\textit{Strong and mid-tier models exhibit distinct vulnerability profiles}: Gemini-3-Pro's largest non-Risk drop is Visual Complexity (35.0pp), while Gemini-3-Flash's is Info Complexity and Reasoning/Calc (both 42.2pp), confirming that multi-dimensional profiling reveals meaningful capability differences.

\section{Discussion: Sim-to-Real Gap}
\label{app:simtoreal}

A natural concern with any benchmark based on generated web environments is the \emph{sim-to-real gap}: do agent capabilities measured on WebForge-Bench transfer to real-world web tasks?

\noindent\textbf{Sources of the gap.} WebForge environments differ from live websites in several ways: (1)~they are self-contained static sites without server-side backends, so tasks involving true server communication (e.g., real API calls, database queries) are not represented; (2)~the visual designs, while professional, are generated rather than produced by dedicated design teams; (3)~the websites have finite scope---a real e-commerce site has millions of products, while a WebForge site has tens to hundreds.

\noindent\textbf{How WebForge mitigates the gap.} Several design choices narrow the gap compared to existing controlled benchmarks: (1)~\textit{Real data and design grounding}---Generation Agent collects real images, product information, and news text from live websites (e.g., Amazon, Booking.com), so the content agents encounter is authentic rather than synthetic. It also references real websites in the same domain as design exemplars rather than inventing pages purely from scratch, making the resulting layouts, visual hierarchy, styling cues, and interaction patterns more consistent with real web interfaces; (2)~\textit{Real-web noise injection}---Refinement Agent injects pop-ups, cookie consent dialogs, and network delays, reproducing the distractions agents face on real websites but that ``sterile'' environments like WebArena omit; (3)~\textit{Anti-cheating mechanisms}---encrypted storage and deceptive codes prevent agents from finding shortcuts that would not exist on real sites; (4)~\textit{Browser-verified solvability}---Validation Agent runs in the same Chromium environment as tested agents, ensuring tasks are genuinely solvable through browser interactions.

\noindent\textbf{Empirical evidence for transferability.} While a formal cross-benchmark correlation study is beyond the scope of this work, the model ranking on WebForge-Bench (Gemini-3-Pro $>$ Claude-4.5-Sonnet $>$ Gemini-3-Flash $>$ GPT-5.2) is broadly consistent with rankings reported on other browser agent benchmarks. Notably, the Browser Use Benchmark\footnote{\url{https://browser-use.com/posts/ai-browser-agent-benchmark}}---an independent, open-source evaluation comprising 100 hard tasks drawn from WebBench, Mind2Web~2, GAIA, and BrowseComp---reports the same top-tier ordering (Gemini-3-Pro $>$ Claude-4.5-Sonnet $>$ GPT-5), corroborating that the capability hierarchy measured on WebForge-Bench generalizes beyond our generated environments. Furthermore, the strong impact of visual input removal (14--16pp drop) mirrors findings from VisualWebArena, confirming that WebForge tasks genuinely test visual understanding capabilities relevant to real-world browsing.

\noindent\textbf{Remaining limitations.} WebForge can simulate simplified versions of certain complex web scenarios---for example, it can mock API responses with static JSON data, emulate basic authentication flows with client-side state, and approximate multi-step transactional workflows using \texttt{localStorage}. However, these approximations have clear boundaries. Tasks requiring \emph{real-time information} (e.g., live stock prices or flight availability that change by the second), \emph{true multi-user interaction} (e.g., collaborative document editing where multiple cursors and conflict resolution are essential), or \emph{persistent server-side state} (e.g., database transactions with rollback semantics, concurrent writes, or cross-session consistency) remain beyond what a static, self-contained web environment can faithfully reproduce. As web agent capabilities advance toward these more complex, stateful scenarios, extending the generation pipeline to support lightweight backend simulation (e.g., embedded serverless functions or ephemeral containers) is an important direction for future work.

\section{Pipeline Walkthrough: End-to-End Demo}
\label{app:demo}

To illustrate the full WebForge pipeline in action, we trace a single task---a \emph{wedding venue booking} in the Consumer Transaction domain at Level~3 difficulty---through all four pipeline stages. This section presents the Plan Agent's two-stage output; subsequent subsections (to follow) will cover the Generation Agent, Refinement Agent, and Validation Agent stages.

\subsection{Plan Agent --- Stage 1: Draft Plan}
\label{app:demo_draft}

The Plan Agent first generates a high-level \emph{draft plan} using Gemini-3-Pro with high-temperature sampling ($T{=}2.0$) for creative diversity. The draft below is the raw output for a wedding venue booking task.

\begin{draftplanbox}[\fcolorbox{draftframe}{draftframe}{\color{white}\sffamily Stage 1} \textsf{Draft Plan Output} \hfill \textsf{\scriptsize Gemini-3-Pro, $T{=}2.0$}]

\textbf{\large Wedding Venue Visual Analytics \& Booking}

\smallskip
\textbf{1.\ Overview}
\begin{itemize}[leftmargin=*,nosep]
\item \textbf{User Query}: ``I need to book the `Grand Estate Gardens' for a wedding in May 2026. Please analyze their availability charts. I need a date where the venue rental price is in the `Standard' or `Economy' tier (indicated by Yellow or Green on their Pricing Heatmap) AND the `White Roses' are in `Peak Bloom' (visible on their Seasonal Flora Chart). Once you identify a valid date that satisfies both visual conditions, book it for 80 guests with the `Premium Catering' add-on. Provide the confirmation code and the final calculated price.''
\item \textbf{Domain}: Consumer Transaction / Service
\item \textbf{Overall Difficulty}: Level 3
\end{itemize}

\smallskip
\textbf{2.\ Difficulty Configuration}

\smallskip
{\scriptsize
\begin{tabularx}{\linewidth}{@{}llX@{}}
\toprule
\textbf{Dimension} & \textbf{Level} & \textbf{Justification} \\
\midrule
Jump Depth & L3 & 7--8 clicks through search, venue, pricing, flora, booking \\
Jump Breadth & L1 & Focused navigation; lists are short (3--5 items) \\
Page Interaction & L2 & Booking form: date, guest count, catering selection \\
Visual Complexity & L3 & Correlating a color-coded heatmap and a line graph \\
Info Complexity & L2 & Chart legends, pricing tiers, catering per-head costs \\
Reasoning/Calc & L3 & Visual intersection + arithmetic (base + 80$\times$rate) \\
Risk Factor & L2 & Transaction with confirmation review step \\
\bottomrule
\end{tabularx}
}

\smallskip
\textbf{3.\ Web Environment Design} --- 7 pages:
\begin{enumerate}[nosep]
\item \textbf{Venue Finder Home} --- Search bar and ``Top Categories.''
\item \textbf{Search Results} --- List of venues including ``Grand Estate Gardens.''
\item \textbf{Venue Dashboard} --- Tabs: Overview, Pricing \& Availability, Gardens \& Flora, Book Now.
\item \textbf{Pricing \& Availability} --- Heatmap image + legend. May 1--14 Red (\$5{,}000), May 15--20 Yellow (\$3{,}500), May 21--31 Green (\$2{,}000).
\item \textbf{Gardens \& Flora} --- Bloom chart. White Roses peak between May 12--18.
\item \textbf{Booking Request} --- Date picker, guest count, catering radio (Standard \$50/pp, Premium \$85/pp, Luxe \$120/pp).
\item \textbf{Booking Confirmation} --- Confirmation code + total cost.
\end{enumerate}

\textbf{Logic Trap}: Valid dates = Yellow/Green $\cap$ Peak Bloom = May 15--18.

\smallskip
\textbf{4.\ Solution Path}
\begin{enumerate}[nosep]
\item Navigate to Home, search ``Grand Estate Gardens.''
\item Click result $\rightarrow$ Venue Dashboard.
\item Pricing tab. \textbf{Visual Step A}: May 15--31 valid (Yellow/Green).
\item Flora tab. \textbf{Visual Step B}: White Rose peak = May 12--18.
\item \textbf{Reasoning}: Intersection $=$ May 15--18. Select any valid date.
\item Book Now. Enter date, 80 guests, Premium (\$85/pp). Submit.
\item Capture confirmation code and total.
\end{enumerate}

\smallskip
\textbf{5.\ Answer}: Mixed (Confirmation Code + Total Cost).
GT Total = \$3{,}500 + 80$\times$\$85 = \textbf{\$10{,}300}. Validate code (\eg, \#WED-9982) and total.

\end{draftplanbox}

\smallskip
\noindent\textbf{Key characteristics of the draft}: The draft captures the core idea---cross-referencing a pricing heatmap and a bloom chart---but contains several simplifications: (1)~only 3 pricing tiers with coarser ranges, (2)~no service fee, (3)~no Saturday preference, (4)~only 7 pages without a review step, (5)~\$85/pp catering, and (6)~ground truth total \$10{,}300.

\subsection{Plan Agent --- Stage 2: Refined Plan}
\label{app:demo_refined}

The Plan Agent then refines the draft using Claude-4.5-Opus with low-temperature sampling ($T{=}1.0$), adding structural detail, difficulty calibration, and quality assurance.

\begin{refinedplanbox}[\fcolorbox{refinedframe}{refinedframe}{\color{white}\sffamily Stage 2} \textsf{Refined Plan Output} \hfill \textsf{\scriptsize Claude-4.5-Opus, $T{=}1.0$}]

\textbf{\large Wedding Venue Availability Analysis \& Booking}

\smallskip
\textbf{1.\ Overview}
\begin{itemize}[leftmargin=*,nosep]
\item \textbf{User Query}: ``I'm planning a wedding at the Grand Estate Gardens in May 2026. I need your help figuring out the best date. On their website, there's a color-coded pricing calendar for May---I only want dates in the Yellow (`Standard') or Green (`Economy') tiers since we're budget-conscious. But I also really want the White Roses to be in full peak bloom for photos, and they have a bloom timeline chart on their Gardens page. Can you cross-reference those two visuals, pick a valid Saturday if possible, and then complete the booking for 80 guests with Premium Catering? I need the confirmation code and total cost when you're done.''
\item \textbf{Domain}: Consumer Transaction / Service
\item \textbf{Overall Difficulty}: Level 3
\end{itemize}

\smallskip
\textbf{2.\ Difficulty Configuration}

\smallskip
{\scriptsize
\begin{tabularx}{\linewidth}{@{}llX@{}}
\toprule
\textbf{Dimension} & \textbf{Level} & \textbf{Justification} \\
\midrule
Jump Depth & L3 & 8 transitions: Home $\to$ Search $\to$ Overview $\to$ Pricing $\to$ Flora $\to$ Book $\to$ Review $\to$ Confirm \\
Jump Breadth & L2 & 5 venues in search; 5 tabs in dashboard; 4 catering options \\
Page Interaction & L2 & 4 form interactions: date, guests, catering, contact name \\
Visual Complexity & L3 & Cross-reference a 4-color heatmap AND a 4-line bloom chart \\
Info Complexity & L2 & Pricing legends, flora descriptions, catering packages, hidden service fee \\
Reasoning/Calc & L3 & Decode heatmap $\to$ dates, read bloom peak, set intersection, Saturday filter, total w/ fee \\
Risk Factor & L2 & Booking transaction with review/confirmation step \\
\bottomrule
\end{tabularx}
}

\smallskip
\textbf{Validation}: 3 dims at L3 + 4 at L2. Meets Level~3 rules ($\geq$2 at L3, $\geq$2 at L2). \checkmark

\smallskip
\textbf{3.\ Web Environment Design} --- 8 pages:
\begin{enumerate}[nosep]
\item \textbf{Venue Finder Home} (\texttt{/}) --- Search bar, featured categories, promo banner (distractor), testimonials.
\item \textbf{Search Results} (\texttt{/search}) --- 5 venue cards. ``Grand Estate Gardens'' is \#1; 4 distractors.
\item \textbf{Venue Overview} (\texttt{/venues/grand-estate-gardens}) --- 5 tabs. \textbf{Callout}: ``\textbf{10\% service fee} applies to all bookings.''
\item \textbf{Pricing \& Availability} (\texttt{/pricing}) --- Heatmap with \textbf{4 tiers}: Red (May 1--10, \$5{,}500), Orange (May 11--14, \$4{,}200), Yellow (May 15--21, \$3{,}200), Green (May 22--31, \$2{,}000). Legend \emph{inside image}. Booked dates: May 3, 9, 24.
\item \textbf{Gardens \& Flora} (\texttt{/flora}) --- 4 flower lines: White Roses peak \textbf{May 13--19}; Lavender all month (distractor); Peonies May 5--12; Hydrangeas May 20--31.
\item \textbf{Book Now} (\texttt{/book}) --- Form: date, guests (50--200), catering (None/Standard \$55/\textbf{Premium \$90}/Luxe \$140), contact, email. Dynamic pricing w/ service fee.
\item \textbf{Review \& Confirm} (\texttt{/book/review}) --- Venue \$3{,}200 + Catering \$7{,}200 + Fee \$1{,}040 = \textbf{\$11{,}440}.
\item \textbf{Confirmation} (\texttt{/confirmation}) --- Code \textbf{GEG-2026-05841}, total \textbf{\$11{,}440.00}.
\end{enumerate}

\smallskip
\textbf{4.\ Solution Path} (17 steps):
\begin{enumerate}[nosep]
\item Navigate to Home, locate search bar.
\item Search ``Grand Estate Gardens'' $\to$ 5 results.
\item Click ``View Details'' on Grand Estate Gardens.
\item Click ``Pricing \& Availability'' tab.
\item \textbf{Visual Analysis A}: Yellow (May 15--21) and Green (May 22--31) are valid.
\item Click ``Gardens \& Flora'' tab.
\item \textbf{Visual Analysis B}: White Roses peak = May 13--19.
\item \textbf{Cross-reference}: May 15--19. Saturday $\to$ \textbf{May 16} (only Saturday).
\item Click ``Book Now.''
\item Date: 2026-05-16 $\to$ Venue \$3{,}200.
\item Guests: 80.
\item ``Premium Plated'' (\$90/pp) $\to$ Catering \$7{,}200.
\item Contact name.
\item Verify: subtotal \$10{,}400 + 10\% fee \$1{,}040 = \$11{,}440.
\item ``Review Booking'' $\to$ Review page.
\item ``Confirm \& Pay Deposit'' $\to$ Confirmation.
\item Read code \textbf{GEG-2026-05841} and total \textbf{\$11{,}440.00}.
\end{enumerate}

\smallskip
\textbf{5.\ Answer Configuration}:
GT: Code = GEG-2026-05841, Total = \$11{,}440.00, Date = 2026-05-16.\\
Full credit: correct code + total + date in May 15--19.\\
75\%: non-Saturday valid date. \quad 50\%: forgot service fee ($\to$ \$10{,}400).

\smallskip
\textbf{6.\ Quality Assurance}: ``Premium Catering'' $\mapsto$ ``Premium Plated'' (intent matching); 10\% fee only in Overview callout; May 2026 starts Friday; 3 distractor flowers.

\end{refinedplanbox}

\smallskip
\noindent\textbf{Key improvements from draft to refined plan.} \cref{tab:plan_diff} summarizes the principal differences. The refined plan transforms a basic 7-page sketch into a detailed 8-page blueprint with richer difficulty calibration, realistic pricing, hidden complexity (service fee), and a narrower solution space (Saturday constraint reduces valid dates from 4 to 1).

\begin{table}[htbp]
\centering
\scriptsize
\caption{\textbf{Draft vs.\ Refined Plan comparison} for the wedding venue booking task.}
\label{tab:plan_diff}
\begin{tabular}{@{}lll@{}}
\toprule
\rowcolor{headercolor}
\textbf{Aspect} & \textbf{Draft Plan} & \textbf{Refined Plan} \\
\midrule
Pricing tiers & 3 (Red/Yellow/Green) & 4 (Red/Orange/Yellow/Green) \\
Venue rental (Yellow) & \$3{,}500 & \$3{,}200 \\
Catering (Premium) & \$85/pp & \$90/pp \\
Service fee & None & 10\% of subtotal \\
Saturday preference & Not mentioned & Explicitly required \\
Valid dates & May 15--18 (4 dates) & May 15--19 (5 dates, 1 Saturday) \\
Total pages & 7 & 8 (added Review step) \\
Jump Breadth & L1 & L2 (5 venues, 5 tabs, 4 catering) \\
Ground truth total & \$10{,}300 & \$11{,}440 \\
Distractor design & Minimal & Promotional banner, sidebar ads, \\
 & & 3 distractor flowers, booked dates \\
\bottomrule
\end{tabular}
\end{table}

\subsection{Generation Agent --- Web Environment Construction}
\label{app:demo_generation}

Given the refined plan, the Generation Agent constructs the complete web environment. A key step is \emph{reference-driven generation}: the agent first browses real websites in the target domain to learn visual styling, interaction patterns, and content tone, then synthesizes a new website that embeds these real-world qualities while implementing the plan's specifications.

\subsubsection{Reference Website Analysis}

For this wedding venue task, the Generation Agent visited 3 real websites and extracted representative visual and interaction priors from them. Below, we visualize these 3 reference-transfer pairs with the clearest evidence.

\paragraph{Representative reference-transfer examples.}

\begin{figure}[htbp]
\centering
\includegraphics[width=0.88\linewidth]{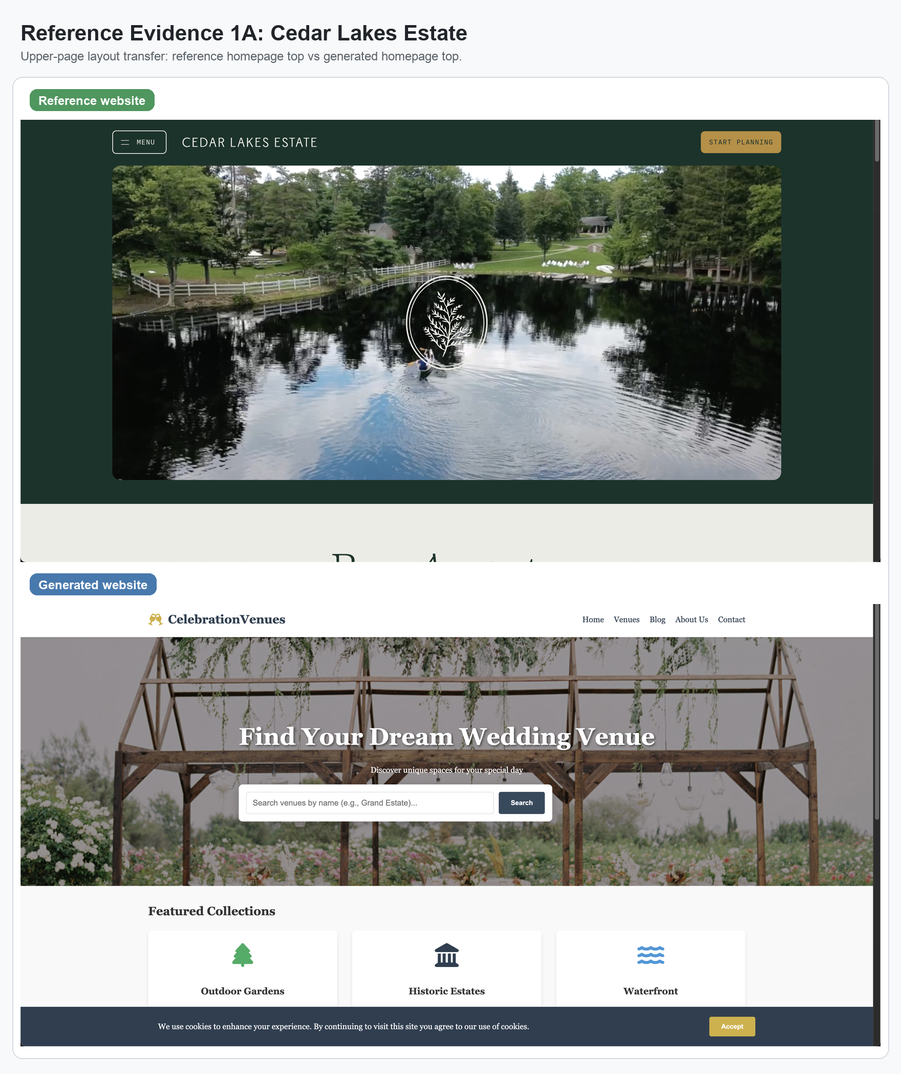}
\caption{\textbf{Cedar Lakes Estate $\rightarrow$ generated homepage (upper-page comparison).} The \emph{top} of the real Cedar Lakes Estate homepage and the \emph{top} of the generated CelebrationVenues homepage exhibit the same high-level homepage prior: a cinematic hero image, restrained luxury branding, and a structure that foregrounds visual atmosphere before denser task information.}
\label{fig:gen_ref_cedar_top}
\end{figure}

\begin{figure}[htbp]
\centering
\includegraphics[width=0.88\linewidth]{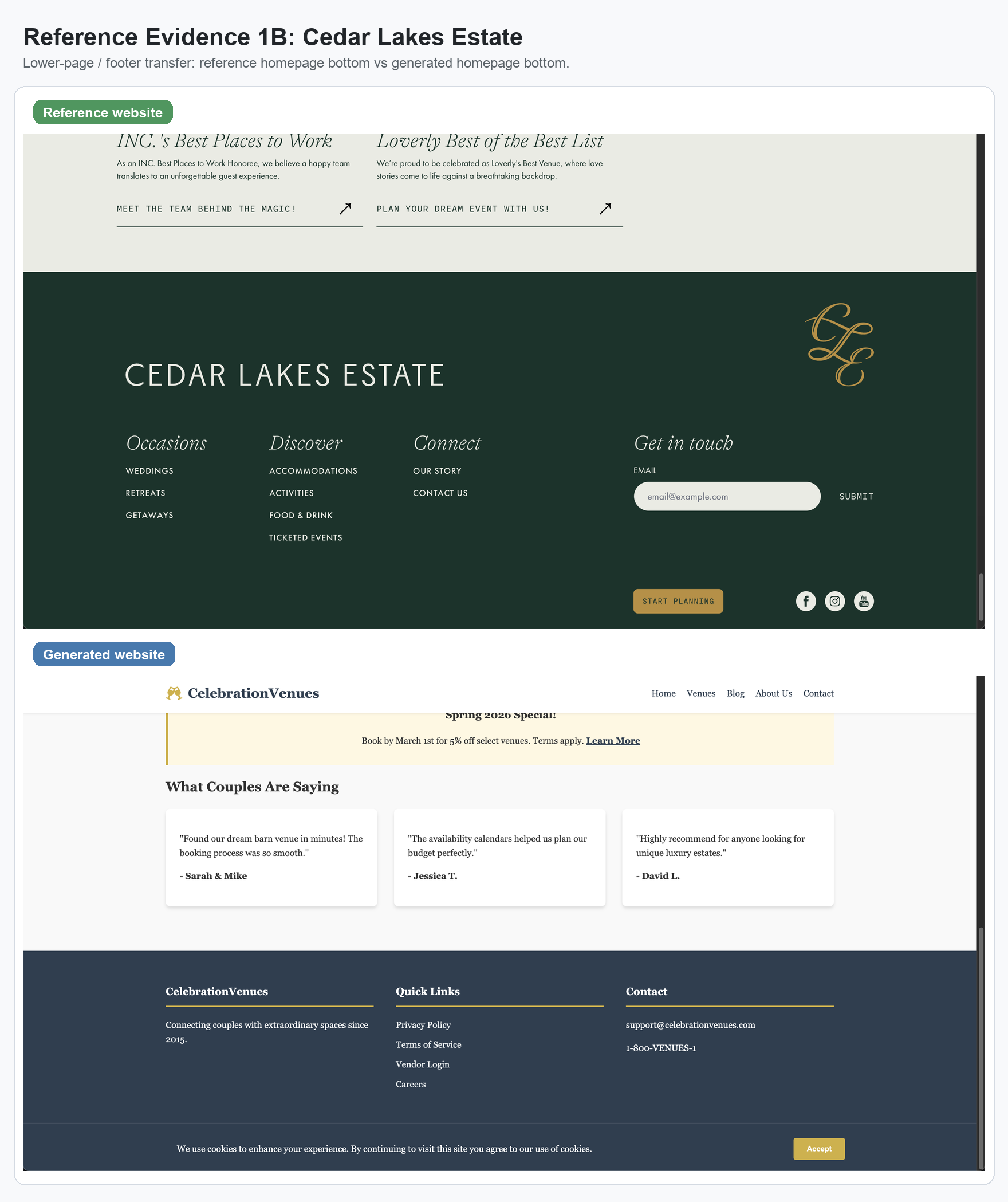}
\caption{\textbf{Cedar Lakes Estate $\rightarrow$ generated homepage (lower-page comparison).} The \emph{bottom} of the reference homepage and the \emph{bottom} of the generated homepage both place substantial visual weight on lower-page promotional content and a prominent footer/contact region, showing that the Generation Agent transferred not only hero styling but also lower-page layout rhythm and brand treatment.}
\label{fig:gen_ref_cedar_bottom}
\end{figure}

\begin{figure}[htbp]
\centering
\includegraphics[width=0.88\linewidth]{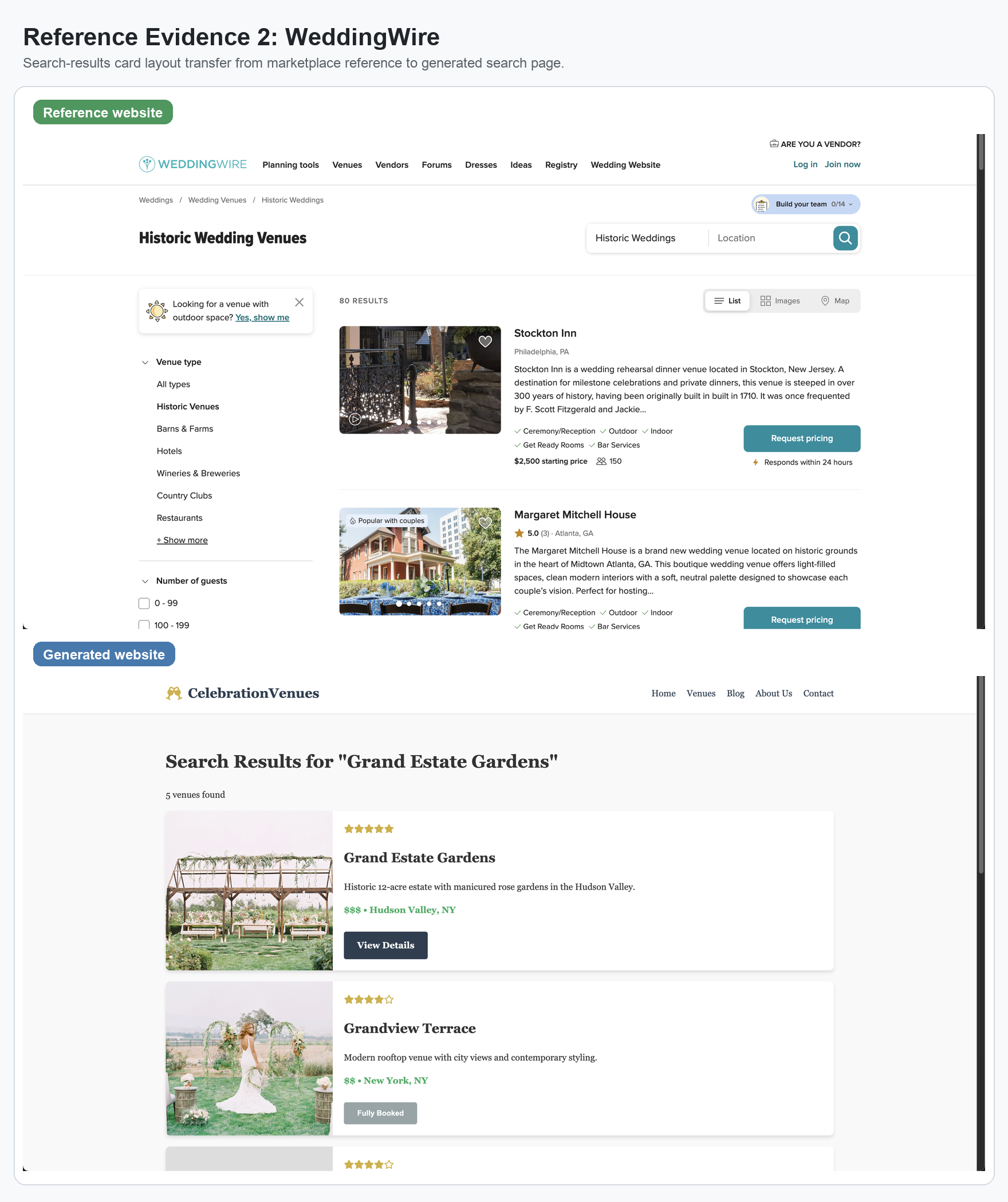}
\caption{\textbf{WeddingWire $\rightarrow$ generated search results page.} The Generation Agent borrows the search-results interaction pattern from WeddingWire: left-aligned thumbnails, stacked result cards, star-rating cues, price-tier markers, and strong venue-selection affordances. This is the clearest evidence that the generated site learned a \emph{retrieval/listing UI pattern} from a real wedding marketplace.}
\label{fig:gen_ref_weddingwire}
\end{figure}

\begin{figure}[htbp]
\centering
\includegraphics[width=0.88\linewidth]{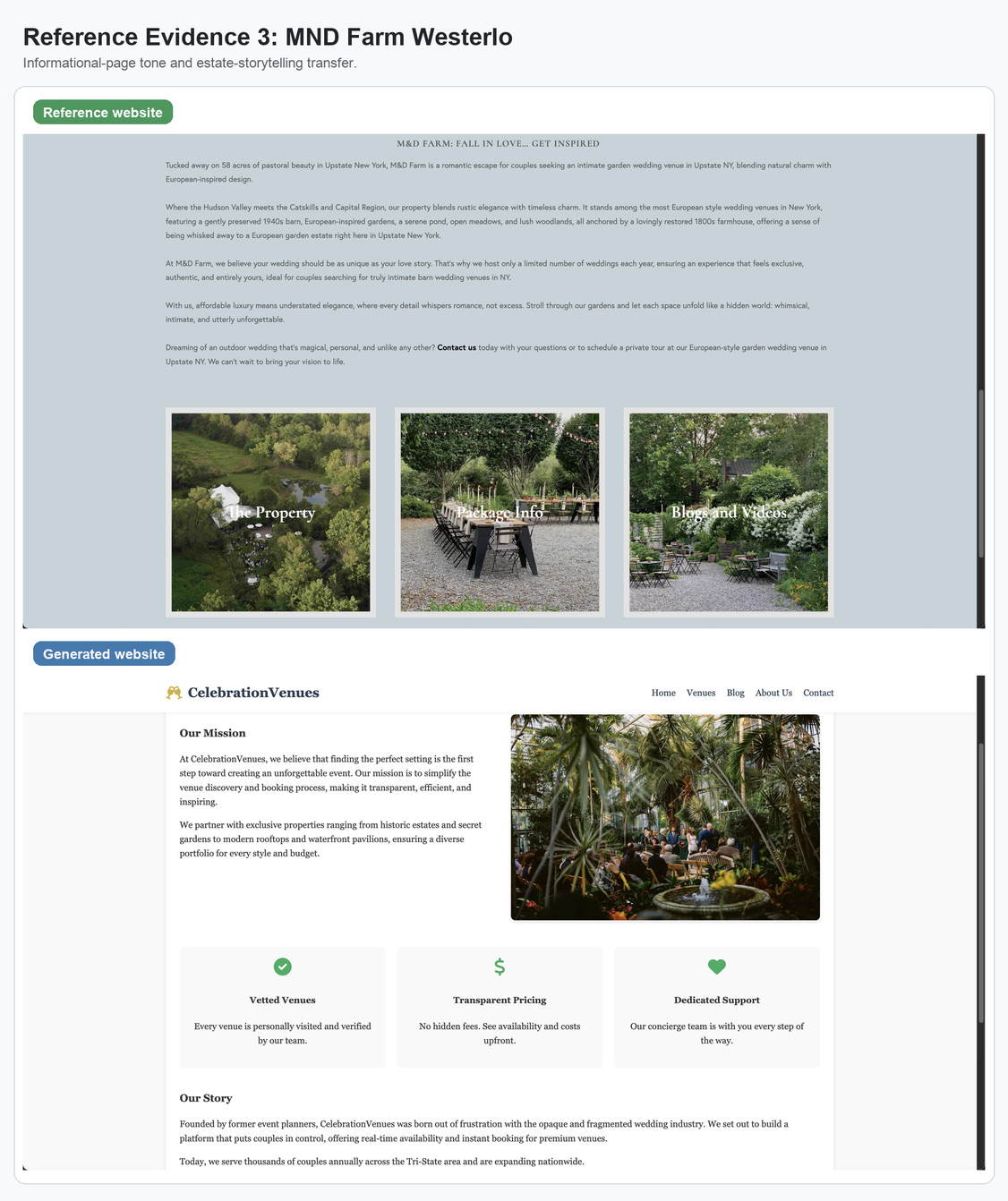}
\caption{\textbf{MND Farm Westerlo $\rightarrow$ generated informational page.} Here the transfer signal is mainly about \emph{editorial tone and venue storytelling}. The real site uses a soft estate narrative with image-backed property presentation; the generated CelebrationVenues page echoes this through a mission/story section, estate-style imagery, and a polished informational layout rather than a purely transactional interface.}
\label{fig:gen_ref_mnd}
\end{figure}

\noindent\textbf{Analysis.} These examples show that the Generation Agent composes its website from \emph{multiple complementary references}: Cedar Lakes Estate contributes global homepage styling and footer rhythm, WeddingWire contributes marketplace search/listing structure, and MND Farm Westerlo contributes narrative tone and informational-page aesthetics. This compositional borrowing behavior is important: the final website is not copied from a single source, but synthesized by combining different real-web priors into a new site that still follows the refined plan.

\smallskip

\subsubsection{Generated Web Environment Overview}

After the reference-driven generation process, the Generation Agent produces a fully functional, self-contained web environment. Table~\ref{tab:gen_output} summarizes the generated artifacts for this wedding venue booking task.

\begin{table}[htbp]
\centering
\caption{\textbf{Generation Agent output summary.} The agent produces a complete web environment comprising HTML pages, visual assets, and supporting code/data files.}
\label{tab:gen_output}
\small
\begin{tabular}{lrl}
\toprule
\textbf{Category} & \textbf{Count} & \textbf{Details} \\
\midrule
HTML pages & 8 & \texttt{index}, \texttt{search}, \texttt{venue\_overview}, \texttt{venue\_pricing}, \\
 & & \texttt{venue\_flora}, \texttt{venue\_book}, \texttt{venue\_review}, \texttt{venue\_confirmation} \\
Image assets & 10 & Scene photos (5), flower photos (3), data charts (2) \\
Code \& data files & 3 & \texttt{style.css}, \texttt{main.js}, \texttt{data.json} \\
\midrule
\textbf{Total files} & \textbf{21} & \\
\bottomrule
\end{tabular}
\end{table}

\begin{figure}[htbp]
\centering
\includegraphics[width=\linewidth]{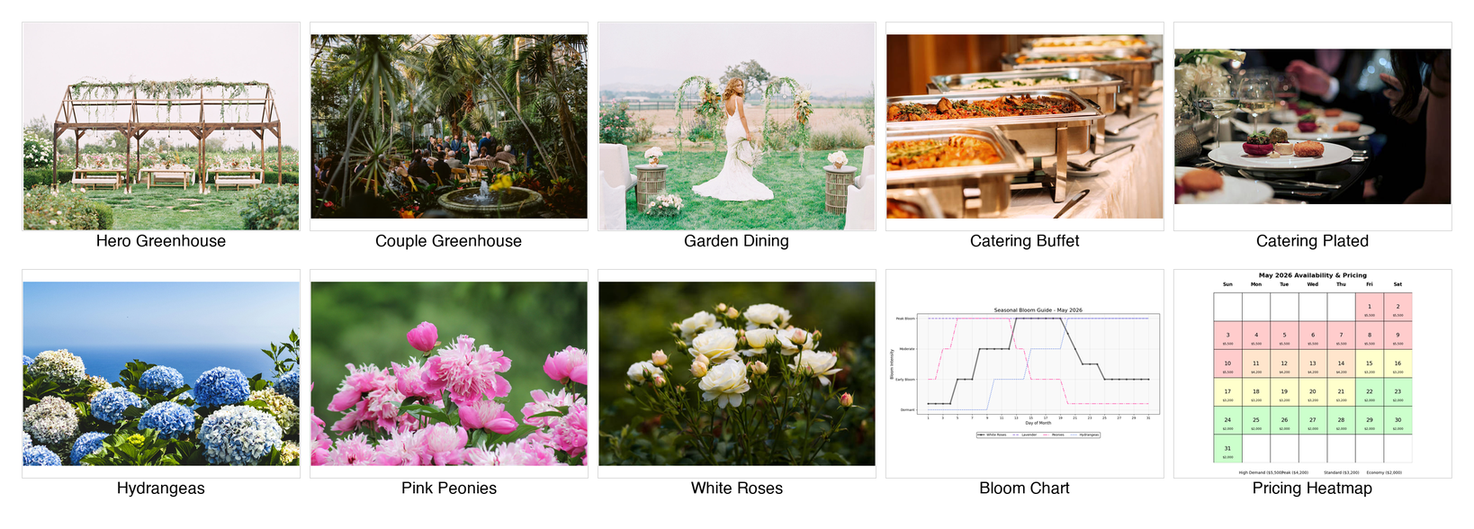}
\caption{\textbf{All 10 image assets produced by the Generation Agent.} Top row: scene and catering photographs used across venue pages. Bottom row: decorative flower images and two data-driven visualizations (a seasonal bloom chart and a pricing/availability heatmap). These images are produced via three tools: Banana (text-to-image generation), web image retrieval, and programmatic Python chart generation.}
\label{fig:gen_image_assets}
\end{figure}

The 10 image assets are produced by three complementary tools available to the Generation Agent: (i)~\textbf{AI image generation} via Banana, a text-to-image generation service, which synthesizes scene-level photographs and decorative visuals from textual prompts (\eg, ``a greenhouse wedding setup with wooden beams and string lights''); (ii)~\textbf{web image retrieval}, which fetches publicly available photographs (\eg, flower close-ups, catering presentations) through search APIs; and (iii)~\textbf{programmatic chart generation}, where the agent writes and executes Python scripts (using libraries such as Matplotlib) to produce data-driven visualizations---in this case, a seasonal bloom guide chart and a pricing/availability calendar heatmap whose content is precisely controlled to encode the task's ground-truth logic. The resulting assets span three categories: venue scene photographs (hero images, dining setups, catering presentations), decorative flower close-ups, and the two programmatically generated charts. Figure~\ref{fig:gen_image_assets} displays all 10 image assets.

\subsection{Refinement Agent --- Rule-Driven Quality Enhancement}
\label{app:demo_refinement}

After the Generation Agent produces the initial web environment (21 files), the Refinement Agent applies a suite of rule-driven transformations to improve realism, accessibility, and task solvability. Table~\ref{tab:ref_diff} summarizes the key changes.

\begin{table}[htbp]
\centering
\caption{\textbf{Before vs.\ After Refinement Agent.} The Refinement Agent expands the website from 21 to 31 files while injecting real-web noise and fixing solvability issues.}
\label{tab:ref_diff}
\small
\begin{tabular}{lcc}
\toprule
\textbf{Metric} & \textbf{Before (Gen.\ Agent)} & \textbf{After (Ref.\ Agent)} \\
\midrule
HTML pages & 8 & 18 (+10 new pages) \\
Image assets & 10 & 10 (unchanged) \\
Code \& data files & 3 & 3 (unchanged) \\
\textbf{Total files} & \textbf{21} & \textbf{31} \\
\midrule
Dead navigation links & Yes (Blog, About, Contact) & Resolved (all links functional) \\
\texttt{alert()} dialogs & Present (blocks DOM parsing) & Replaced with inline errors \\
Real-web noise & None & Cookie banner + stochastic popup \\
Generic venue pages & Missing & ``Content Unavailable'' placeholder \\
\bottomrule
\end{tabular}
\end{table}

\noindent We highlight four representative improvements below, each with visual evidence.

\subsubsection{Navigation Completeness: Resolving Dead Links}

The Generation Agent focused on the 8 pages directly required by the task plan, leaving global navigation links (Blog, About Us, Contact, etc.) non-functional. Clicking them had no effect. The Refinement Agent then created 10 additional pages spanning informational content, account and policy routes, and venue-supporting views, and wired all navigation links to point to them. This makes the website feel like a complete, coherent system rather than a collection of disconnected pages, and prevents browser agents from encountering dead-end states when exploring the site.

\begin{figure}[htbp]
\centering
\includegraphics[width=0.88\linewidth]{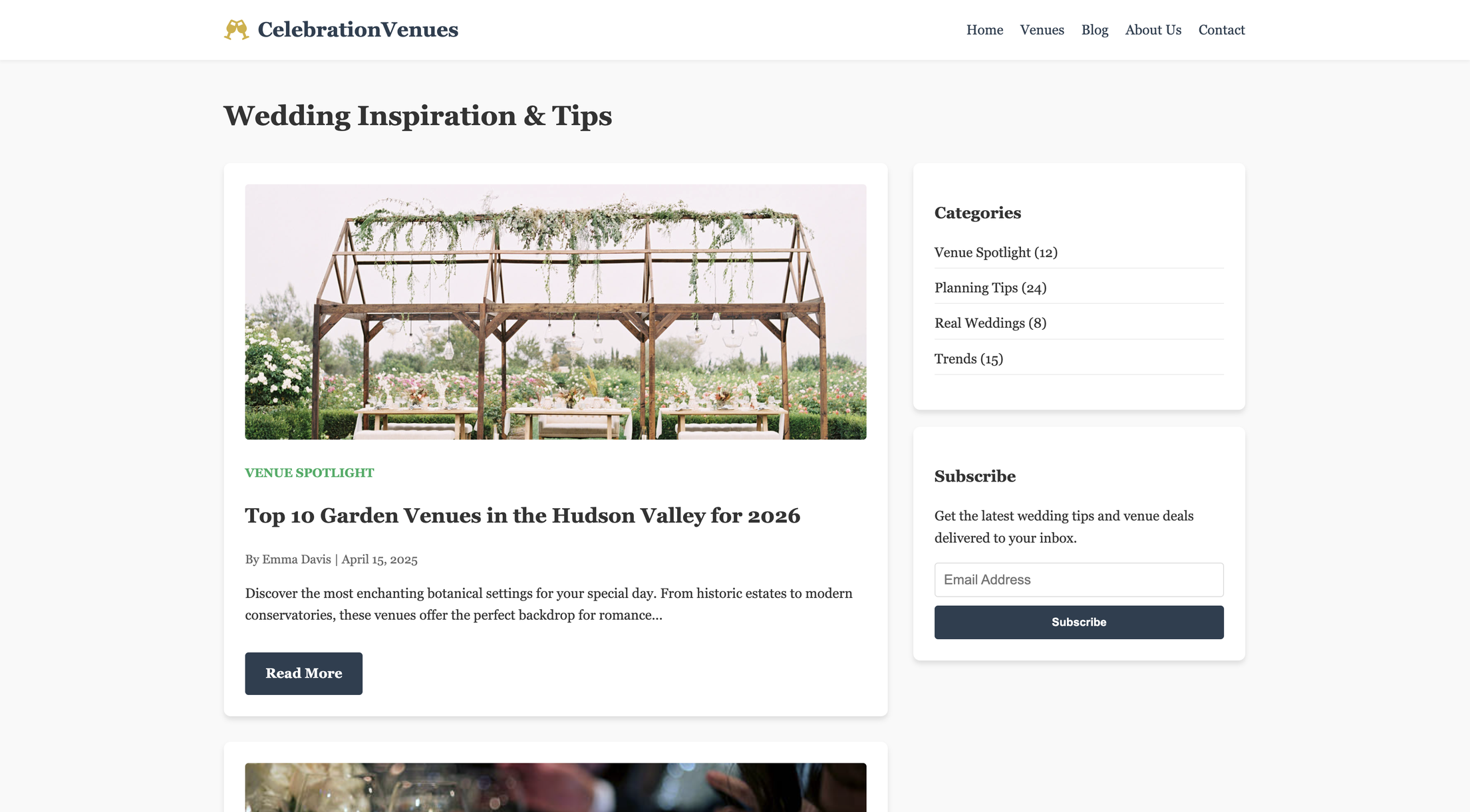}
\caption{\textbf{Blog page created by the Refinement Agent.} The Generation Agent's navigation bar included a ``Blog'' link that led nowhere. The Refinement Agent created a fully functional blog page with categorized articles (``Outdoor,'' ``Planning,'' ``Inspiration''), a sidebar with category filters and a newsletter subscription widget, and consistent site-wide navigation. Similar pages were created for About Us, Contact, Careers, Login, Register, Privacy Policy, and Terms of Service.}
\label{fig:ref_blog}
\end{figure}

Figure~\ref{fig:ref_blog} shows the newly created Blog page as a representative example. The page features categorized wedding inspiration articles, a sidebar with categories and a newsletter subscription widget, and consistent header/footer navigation---all generated by the Refinement Agent to fill a previously empty link.

\subsubsection{Solvability Fix: Replacing \texttt{alert()} with Inline Error Messages}

The Generation Agent used JavaScript \texttt{alert()} dialogs for form validation (e.g., ``Please enter a contact name.''). While visually functional for human users, native \texttt{alert()} dialogs \textbf{block DOM parsing and JavaScript execution}, creating a critical problem for browser-based agents: the agent cannot dismiss the dialog programmatically in many automation frameworks, causing the task to become unsolvable. The Refinement Agent replaced all \texttt{alert()} calls with inline DOM-based error messages that are both human-readable and agent-accessible.

\begin{figure}[htbp]
\centering
\begin{minipage}[t]{0.48\linewidth}
\centering
\includegraphics[width=\linewidth]{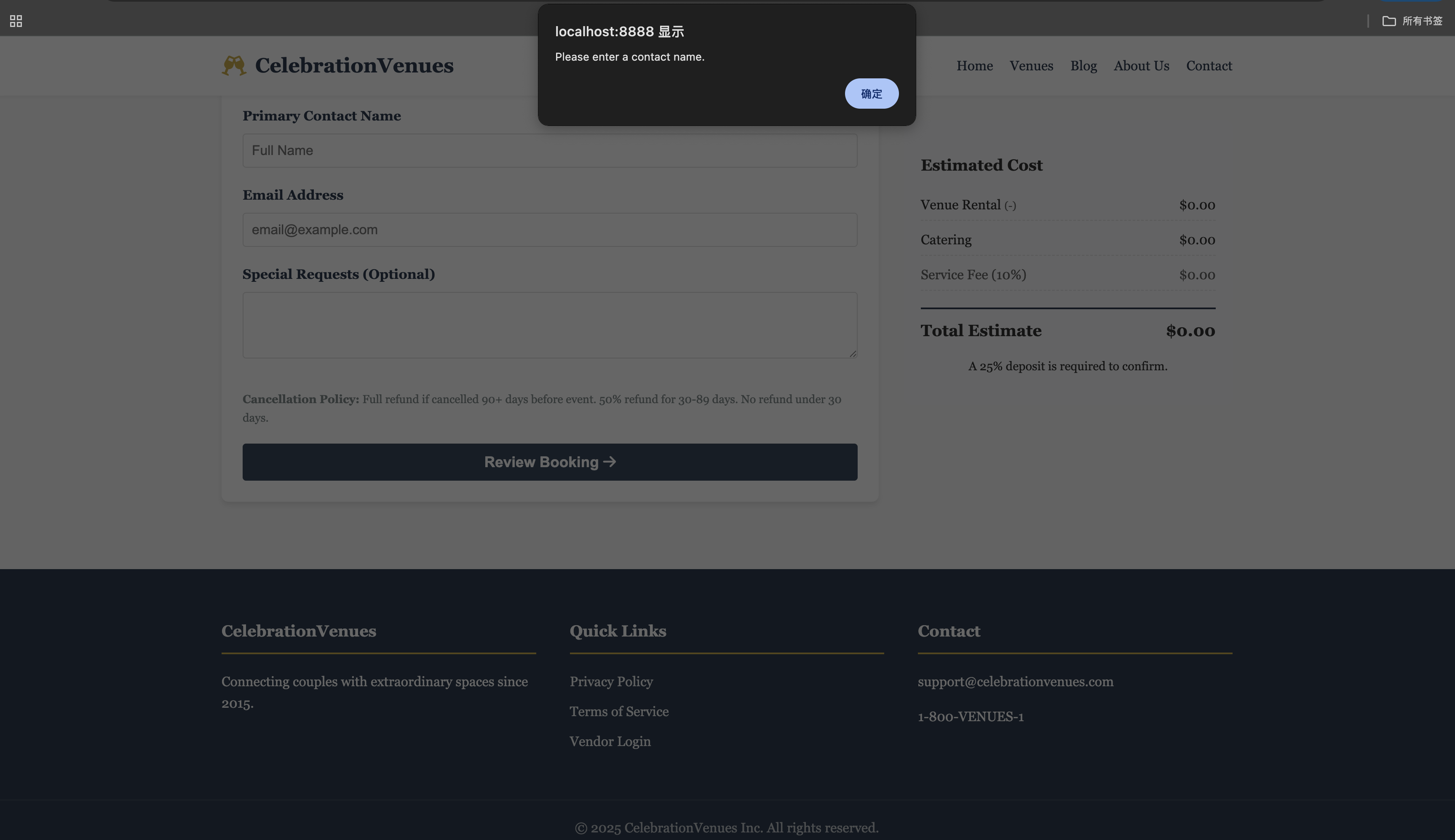}
\end{minipage}
\hfill
\begin{minipage}[t]{0.48\linewidth}
\centering
\includegraphics[width=\linewidth]{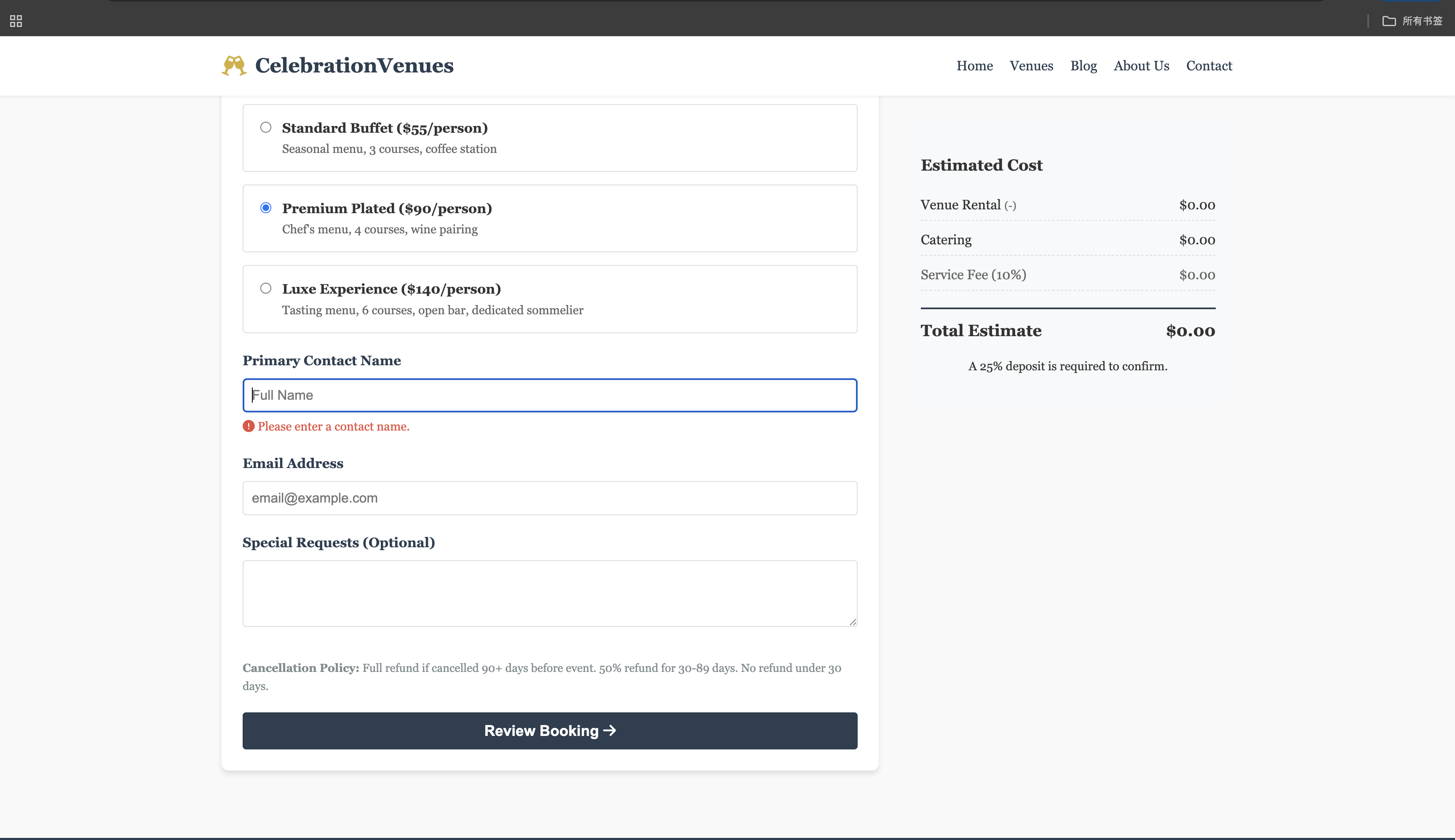}
\end{minipage}
\caption{\textbf{Alert dialog replacement.} \emph{Left}: The Generation Agent's \texttt{alert()} dialog blocks DOM parsing, making the page unresponsive to browser agents. \emph{Right}: The Refinement Agent's inline error message (``$\oslash$ Please enter a contact name.'') appears as styled DOM content below the form field, maintaining full page interactivity and ensuring task solvability for automated agents.}
\label{fig:ref_alert}
\end{figure}

Figure~\ref{fig:ref_alert} shows the before-and-after comparison. The native browser alert (left) blocks all page interaction until manually dismissed; the inline error message (right) appears as styled red text below the input field, preserving full DOM interactivity.

\subsubsection{Real-Web Noise Injection: Stochastic Popup and Cookie Banner}

To simulate the distractions encountered on real websites, the Refinement Agent injects two types of interruptions:

\begin{itemize}[nosep]
\item \textbf{Cookie consent banner}: Appears 1 second after page load on every page, asking users to accept cookies. Persisted via \texttt{localStorage} so it only appears once per session.
\item \textbf{``Schedule a Private Tour'' popup}: A promotional overlay that appears after a \textbf{stochastic delay of 5--15 seconds} (uniformly sampled via \texttt{5000 + Math.random() * 10000} ms). The popup offers a ``10\% discount voucher'' and includes a ``Book Tour Now'' call-to-action and a ``No thanks'' dismissal link. Once dismissed, it is suppressed for future visits via \texttt{localStorage}.
\end{itemize}

These noise elements serve two purposes: (1)~they test agents' ability to handle unexpected overlays that may occlude task-relevant content, and (2)~they make the web environment more realistic, narrowing the sim-to-real gap discussed in Appendix~\ref{app:simtoreal}. Figure~\ref{fig:ref_popup} shows both interruptions simultaneously visible on the page.

\begin{figure}[htbp]
\centering
\includegraphics[width=0.88\linewidth]{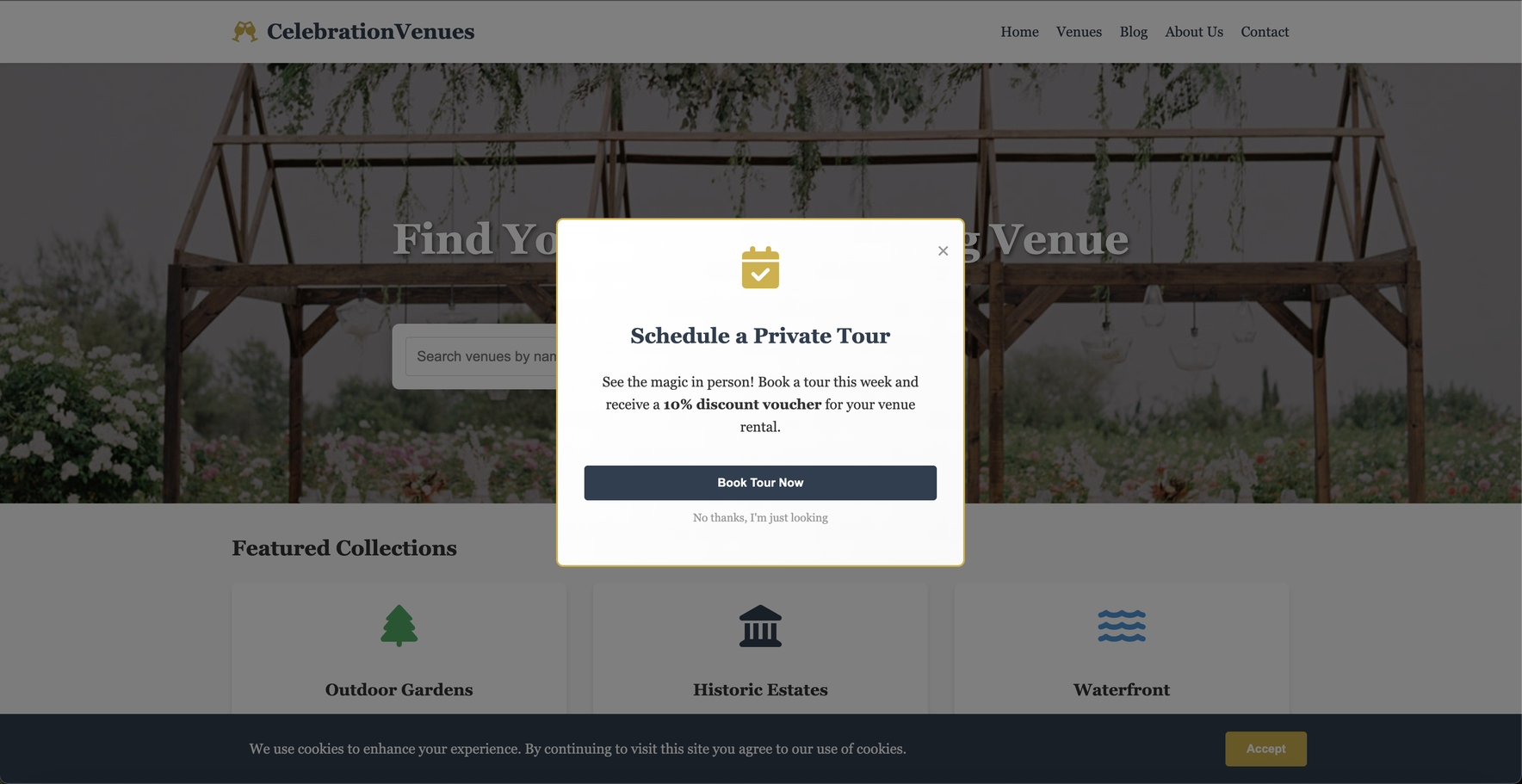}
\caption{\textbf{Real-web noise injection.} The ``Schedule a Private Tour'' promotional popup (center) appears after a stochastic 5--15s delay, while the cookie consent banner (bottom) appears 1s after page load. Together they simulate the real-world browsing distractions that agents must handle gracefully.}
\label{fig:ref_popup}
\end{figure}

\subsubsection{Dead Link Resolution: Generic Venue Placeholder}

The search results page lists 5 venues, but only ``Grand Estate Gardens'' (the target venue) has a fully implemented detail page. The Generation Agent left the ``View Details'' buttons for the other 4 venues as dead links. The Refinement Agent addresses this by creating a generic \texttt{venue\_generic.html} placeholder page that displays a ``Content Unavailable'' notice with ``Contact Support'' and ``Back to Search'' action buttons, as well as a ``Similar Venues You Might Like'' section to maintain site coherence. All non-target venue links now point to this page.

This improvement prevents agents from encountering broken navigation when exploring non-target venues---a realistic scenario since agents may click on distractor venues before finding the correct one. The placeholder page gently redirects agents back to productive navigation paths.

\begin{figure}[htbp]
\centering
\includegraphics[width=0.88\linewidth]{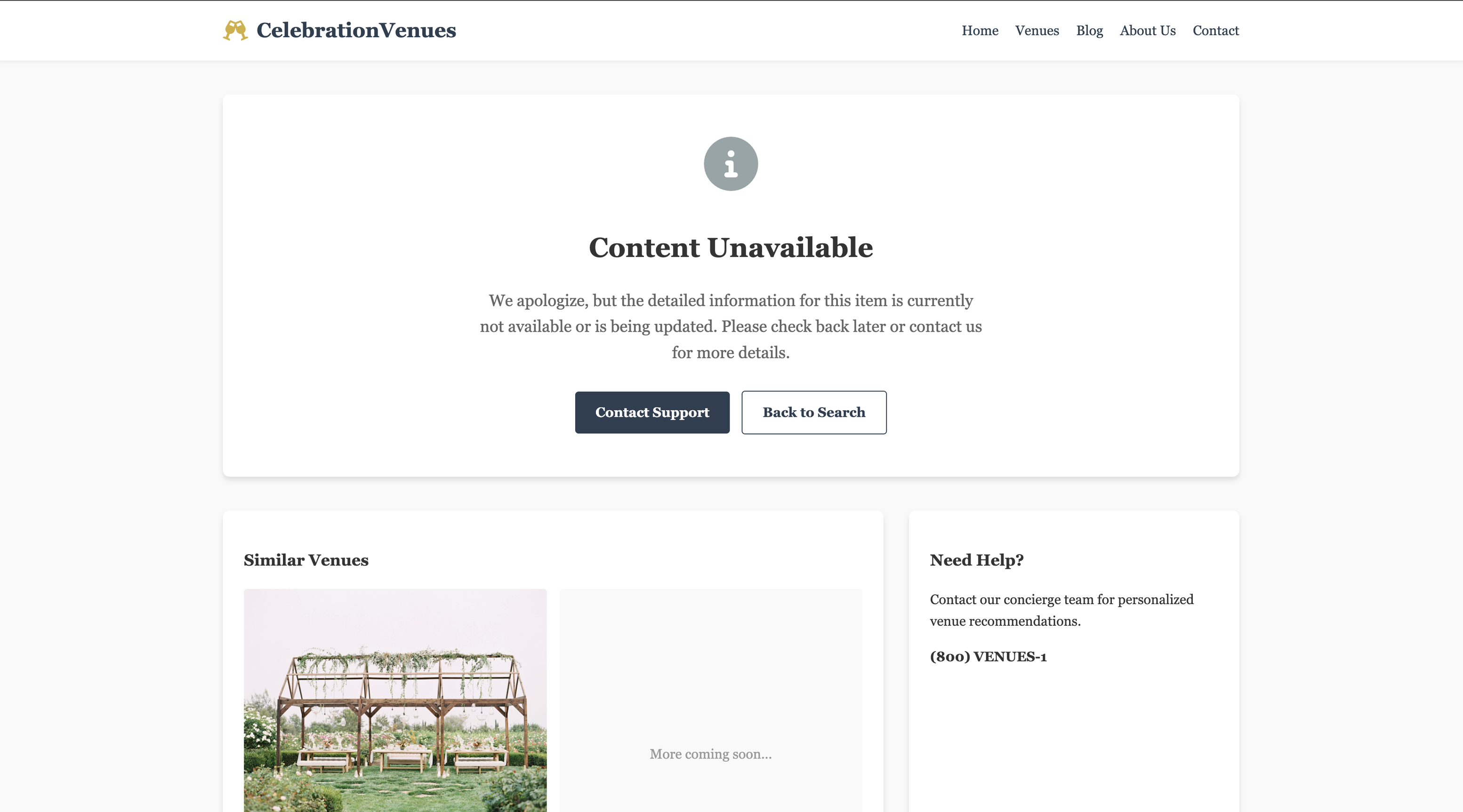}
\caption{\textbf{Generic venue placeholder page.} When agents click ``View Details'' on non-target venues, they now see this ``Content Unavailable'' page instead of encountering a dead link. The page provides clear navigation options (``Contact Support,'' ``Back to Search'') and suggests similar venues, maintaining a realistic user experience while guiding agents back to productive paths.}
\label{fig:ref_generic_venue}
\end{figure}

\subsubsection{Additional Refinements}

Beyond the four improvements visualized above, the Refinement Agent also applies several other enhancements:
\begin{itemize}[nosep]
\item \textbf{Login and Registration pages}: Added \texttt{login.html} and \texttt{register.html} with form validation, providing complete user authentication flows even though the task does not require logging in.
\item \textbf{Legal and policy pages}: Added \texttt{privacy.html}, \texttt{terms.html}, and \texttt{careers.html} to populate footer links, creating a professional and complete website structure.
\item \textbf{Venue gallery page}: Added \texttt{venue\_gallery.html} as a visual browsing alternative to the list-based search page.
\item \textbf{Search logic enhancement}: Updated the search functionality so that queries not matching the target venue still display results (the 5-venue list), rather than showing an empty page or error.
\end{itemize}

\noindent Together, these refinements transform a minimal task-focused prototype into a realistic, fully navigable web environment that tests agents under conditions closer to real-world browsing.

\subsection{Anti-Cheating: Encryption and Deceptive Codes}
\label{app:demo_anticheating}

A critical design requirement for browser benchmarks is preventing agents from bypassing the intended task by directly reading ground-truth answers from the page source or data files. WebForge implements a two-layer anti-cheating mechanism: \textbf{Base64 encryption} of all sensitive values, and \textbf{deceptive codes} that return plausible but incorrect answers for common mistake patterns.

\subsubsection{Encrypted Data Storage}

All task-critical values---confirmation codes, correct totals, correct dates---are stored as Base64-encoded strings in a separate \texttt{data.json} file, never as plaintext in the HTML or JavaScript source. For the wedding venue task, the data file contains:

\begin{anticheatingbox}[\fcolorbox{anticheatingframe}{anticheatingframe}{\color{white}\sffamily Anti-Cheat} \textsf{Encrypted Answer-Bearing Fields in \texttt{data.json}}]
\begin{verbatim}
{
  "ground_truth": {
    "confirmation_code": "R0VHLTIwMjYtMDU4NDE=",
    "total_cost": "MTE0NDAuMDA=",
    "correct_date": "MjAyNi0wNS0xNg=="
  },
  "deceptive_codes": {
    "wrong_date_generic": "R0VHLTIwMjYtMDUyOTQ=",
    "wrong_catering": "R0VHLTIwMjYtMDU5OTE=",
    "wrong_guests": "R0VHLTIwMjYtMDUxMTg=",
    "valid_non_saturday": "R0VHLTIwMjYtMDU4NDI="
  }
}
\end{verbatim}
\end{anticheatingbox}

\noindent In the actual website, an agent inspecting the source code sees only opaque Base64 strings rather than the plaintext answers. For readability, we also show the decoded values below; these plaintext strings are presented here only for explanation and are not exposed directly in the web environment. WebForge encrypts the \emph{answer-bearing fields} (correct answer and deceptive confirmation codes), rather than every public task constraint.

\begin{anticheatingbox}[\fcolorbox{anticheatingframe}{anticheatingframe}{\color{white}\sffamily Anti-Cheat} \textsf{Decoded Values}]
\begin{verbatim}
{
  "decoded_ground_truth": {
    "confirmation_code": "GEG-2026-05841",
    "total_cost": "11440.00",
    "correct_date": "2026-05-16"
  },
  "decoded_deceptive_codes": {
    "wrong_date_generic": "GEG-2026-05294",
    "wrong_catering": "GEG-2026-05991",
    "wrong_guests": "GEG-2026-05118",
    "valid_non_saturday": "GEG-2026-05842"
  }
}
\end{verbatim}
\end{anticheatingbox}

\subsubsection{Deceptive Code Mechanism}

Rather than returning a generic error for incorrect bookings, the system returns \emph{plausible but wrong} confirmation codes tailored to specific mistake patterns. We do not reproduce the full decision logic from \texttt{main.js} here; the key point for the benchmark is the mapping between mistake types and the codes returned to the agent. Each deceptive code follows the same format (\texttt{GEG-2026-XXXXX}) as the real code, so an agent cannot distinguish correct from incorrect by format alone. Table~\ref{tab:deceptive} shows the full mapping.

\begin{table}[htbp]
\centering
\small
\caption{\textbf{Deceptive code mapping.} Each common mistake pattern returns a unique, plausible-looking confirmation code. Only the fully correct booking yields the ground-truth code.}
\label{tab:deceptive}
\begin{tabular}{@{}llc@{}}
\toprule
\rowcolor{headercolor}
\textbf{Condition} & \textbf{Code Returned} & \textbf{Correct?} \\
\midrule
Correct (May 16, 80 guests, Premium) & GEG-2026-05841 & \checkmark \\
Valid date but not Saturday (May 15/17/18/19) & GEG-2026-05842 & $\times$ \\
Wrong catering selection & GEG-2026-05991 & $\times$ \\
Wrong guest count & GEG-2026-05118 & $\times$ \\
Invalid date (outside May 15--19) & GEG-2026-05294 & $\times$ \\
\bottomrule
\end{tabular}
\end{table}

\noindent\textbf{Why this matters.} This design has three important consequences: (1)~agents \emph{must} complete the full multi-step workflow to obtain any confirmation code, preventing source-code shortcuts; (2)~agents that make partial mistakes (e.g., choosing the right date range but missing the Saturday constraint) receive a code that \emph{looks valid} but is wrong, testing whether the agent truly followed all task constraints; (3)~the deceptive codes enable fine-grained error analysis---by examining which deceptive code an agent received, we can diagnose the specific reasoning failure (wrong catering, wrong date, etc.) rather than just recording a binary pass/fail.

\subsection{Validation Agent --- Browser-Verified Solvability}
\label{app:demo_validation}

The final pipeline stage deploys the Validation Agent---a browser-based agent running in the same Chromium environment as the models under evaluation---to verify that the task is genuinely solvable through standard browser interactions. For this wedding venue task, the Validation Agent completed the task in \textbf{24 steps} over approximately 6 minutes, successfully obtaining the correct confirmation code (\texttt{GEG-2026-05841}) and total cost (\$11,440.00). The complete execution trace is shown below, preserving the agent's original observation, reasoning, and action outputs.

\begin{validationbox}[\fcolorbox{validationframe}{validationframe}{\color{white}\sffamily Validation} \textsf{Full Execution Trace (24 Steps)} \hfill \textsf{\scriptsize Chromium Browser Agent}]
\scriptsize
\begin{verbatim}
Step 1
  Observation: I am on the homepage of CelebrationVenues. I see a
    search bar.
  Reasoning:   I need to find the 'Grand Estate Gardens' venue as the
    first step of the task. I will use the search bar to find it.
  Action:      Input 'Grand Estate Gardens' into search field at
    index 6 | Click search button at index 7

Step 2
  Observation: A 'Schedule a Private Tour' promotional popup is
    currently visible, covering the main content. Index 0 is the
    close button ('x').
  Reasoning:   The popup is blocking interaction with the main page
    elements like the search bar. I must dismiss it to continue with
    the search for 'Grand Estate Gardens'.
  Action:      Click the close button 'x' at index 0 to dismiss the
    promotional popup.

Step 3
  Observation: The popup is gone. I can see the search input field at
    index 6 and the search button at index 7. There is also a cookie
    consent banner at the bottom with an 'Accept' button at index 11.
  Reasoning:   I can now proceed with searching for the venue. I will
    first input the search term and then click search. I'll also
    accept the cookies to clear the UI.
  Action:      Input 'Grand Estate Gardens' into the search field at
    index 6 and click the 'Search' button at index 7.

Step 4
  Observation: I am on the search results page. I see 'Grand Estate
    Gardens' as the first result. The 'View Details' button for it is
    at index 6.
  Reasoning:   The next step in the solution is to click on the venue
    result to view the Overview dashboard. I will click the 'View
    Details' button for Grand Estate Gardens.
  Action:      Click 'View Details' for Grand Estate Gardens at
    index 6.

Step 5
  Observation: I am on the Overview page for Grand Estate Gardens. I
    see tabs for Overview, Pricing & Availability, Gardens & Flora,
    Gallery, and Book Now. I also notice a mention of a 10% service
    fee in the Overview page content.
  Reasoning:   I need to analyze the May 2026 heatmap to identify
    valid pricing tiers. I will navigate to the 'Pricing &
    Availability' tab.
  Action:      Click the 'Pricing & Availability' tab at index 7.

Step 6
  Observation: I am on the Pricing & Availability page. I see the
    'May 2026 Availability & Pricing' heatmap.
    - May 1-10: Red (High Demand)
    - May 11-14: Orange (Peak)
    - May 15-21: Yellow (Standard)
    - May 22-31: Green (Economy)
    The valid ranges for budget (Yellow/Green) are May 15-31.
  Reasoning:   I have identified the budget-friendly dates (May
    15-31). Now I need to find the peak bloom period for White Roses
    to cross-reference with these dates. I will go to the 'Gardens &
    Flora' page.
  Action:      Click the 'Gardens & Flora' tab at index 8.

Step 7
  Observation: I am on the Gardens & Flora page. I see the title
    'Seasonal Bloom Guide - May 2026' and the top of a chart. I need
    to see the full chart.
  Reasoning:   I need to find the peak bloom period for White Roses.
    The chart is partially visible, so I'll scroll down to see it
    clearly and determine the dates. Then I will cross-reference with
    the valid pricing dates (May 15-31).
  Action:      Scroll down to view the bloom chart and determine the
    peak bloom period for White Roses.

Step 8
  Observation: I can see the bloom chart now.
    - White Roses (solid dark line) reach Peak Bloom: May 13-19.
    - Intersection with budget dates (May 15-31) is May 15-19.
    - Saturdays in May 2026: May 2, 9, 16, 23, 30.
    - The only Saturday in the intersection is May 16, 2026.
    May 16 is in the Yellow Tier ($3,200).
  Reasoning:   I have identified the target date: May 16, 2026. This
    date satisfies the budget constraint (Yellow Tier), the flower
    constraint (White Roses in Peak Bloom), and the user's preference
    for a Saturday. Now I will proceed to the booking page.
  Action:      Click the 'Book Now' tab at index 10.

Step 9
  Observation: I am on the Book Now page. I see the 'Event Date'
    input field and 'Number of Guests' field. I need to scroll down
    to see the rest of the form, including catering options and
    contact details.
  Reasoning:   I need to fill out the form with: Date: 2026-05-16,
    Guests: 80, Catering: Premium Plated, Name: Sarah Jenkins,
    Email: sarah.j@example.com. I'll scroll down to access all
    fields.
  Action:      Scroll down to fill out the booking form.

Step 10
  Observation: I can see the form fields.
    - Event Date: index 12
    - Number of Guests: index 14
    - Catering Options: Premium Plated is at index 21.
    I still need to fill in the primary contact name and email.
  Reasoning:   I will fill out the first part of the form now. I'll
    enter the date, guest count, and select the catering package.
    Then I'll scroll down to finish the contact details.
  Action:      Input '2026-05-16' into index 12 | Input '80' into
    index 14 | Click 'Premium Plated' at index 21 | Scroll down to
    fill contact info.

Step 11
  Observation: I'm at the bottom of the booking form.
    - Primary Contact Name: index 15
    - Email Address: index 17
    - Review Booking button: index 20
    The Estimated Cost summary currently shows Venue Rental as
    'Unavailable/Unknown'. This might update after clicking Review.
  Reasoning:   I will enter the contact name and email, then click
    'Review Booking' to see the final cost breakdown. The date
    '2026-05-16' is a Saturday in May 2026 and matches the Yellow
    Tier ($3,200). 80 guests x $90 catering = $7,200. Subtotal
    $10,400. 10% Service Fee $1,040. Total $11,440.
  Action:      Input 'Sarah Jenkins' into index 15 | Input
    'sarah.j@example.com' into index 17 | Click 'Review Booking'
    button at index 20.

Step 12  [ERROR DETECTED]
  Observation: I am on the Review page. The date is incorrectly
    displayed as '0516-02-02' and the Venue Rental is '$0'. The
    total cost is '$7,920.00', which is incorrect compared to the
    expected '$11,440.00'.
  Reasoning:   The date was not correctly processed by the booking
    form, leading to a zero venue rental and a wrong total. I will
    go back and re-enter the date using the format '05/16/2026' as
    suggested by the placeholder in the screenshot.
  Action:      Go back to the 'Book Now' page using the browser back
    button. Then try to input the date in 'mm/dd/yyyy' format.

Step 13
  Observation: I am back on the Book Now page (scrolled down). I see
    the contact details are still filled. I need to scroll up to fix
    the date.
  Reasoning:   I will fix the date and re-submit the form. I'll use
    the 'mm/dd/yyyy' format this time.
  Action:      Scroll up to find the date field at index 12. Input
    '05/16/2026' and then click 'Review Booking' at index 20.

Step 14
  Observation: I am on the Book Now page. The date field (index 12)
    currently says '02/02/0516'. I need to correct this.
  Reasoning:   I will fix the date to the correct '05/16/2026'
    format to ensure the venue rental is calculated. Then I will
    submit the form again.
  Action:      Clear the date field at index 12 and input
    '05/16/2026'. Then scroll down and click 'Review Booking'.

Step 15
  Observation: The browser state says I am on the Review page, but
    the image shows the booking form. The interactive elements are
    missing the form fields. I might be in an inconsistent state.
  Reasoning:   I need to be on the 'Book Now' page to enter the
    correct date. I will navigate to the URL directly to be certain.
  Action:      Navigate to the 'Book Now' page URL directly to
    ensure I'm on the correct page.

Step 16
  Observation: I'm on the Book Now page. The date field at index 12
    is visible. The cost estimate on the right still shows the old
    (incorrect) state. I will re-enter everything.
  Reasoning:   I will enter the date using the YYYY-MM-DD format
    first, as it's more standard for HTML date inputs.
  Action:      Input '2026-05-16' into index 12 | Scroll down and
    input guest count and catering.

Step 17
  Observation: The date input still shows '02/02/0516'. My previous
    input was not successful or was overwritten.
  Reasoning:   I will try the MM/DD/YYYY format this time as it's
    what's shown in the input. I'll also ensure all other fields are
    correct before clicking review.
  Action:      Scroll up and input '05/16/2026' into index 6. Then
    scroll down and click 'Review Booking'.

Step 18
  Observation: I am on the Book Now page. The Venue Rental still
    shows '$0' and 'Unavailable/Unknown'. The date input at the top
    is likely the cause. I've tried multiple formats.
  Reasoning:   The automated input of the date seems to be failing
    to trigger the price update. I will try to input the date as
    '2026-05-16' again but this time I will click on another field
    after inputting to trigger any 'onchange' events.
  Action:      Scroll up. Manually click the date input field and
    type '05162026'. Then click 'Review Booking'.

Step 19
  Observation: I'm at the top of the form. Date input is index 12.
    Guest input is index 14. Catering 'Premium Plated' is selected.
    Contact details are likely still filled further down.
  Reasoning:   I will try a very deliberate interaction to ensure
    the date is accepted and the price is calculated. I'll click,
    type, and then click elsewhere.
  Action:      Click on the date input at index 12. Type
    '05/16/2026'. Then click on the guest count field at index 14
    to trigger any change event.

Step 20  [SUCCESS - Price updated correctly]
  Observation: Great! The Estimated Cost now shows the correct
    values:
    - Venue Rental (Standard): $3,200
    - Catering: $7,200
    - Service Fee (10%): $1,040.00
    - Total Estimate: $11,440.00
    This matches my calculation.
  Reasoning:   The cost is now correct. I'll scroll down to ensure
    the contact details are still there and then click 'Review
    Booking'. After that, I'll confirm the booking.
  Action:      Scroll down and click 'Review Booking'. Then on the
    Review page, click 'Confirm & Pay Deposit'.

Step 21
  Observation: The page hasn't transitioned yet. I'm still on the
    Book Now page. All fields are filled correctly and the estimate
    is right. Index 15 is the 'Review Booking' button.
  Reasoning:   The previous click might have failed. I'll click the
    button again.
  Action:      Click 'Review Booking' at index 15.

Step 22
  Observation: I am on the Review page. The summary is correct:
    - Venue: Grand Estate Gardens
    - Date: 2026-05-16
    - Guests: 80
    - Catering: Premium
    - Venue Rental: $3,200
    - Catering Total: $7,200
    - Service Fee (10%): $1,040.00
    - Total Cost: $11,440.00
    I need to find the confirm button.
  Reasoning:   Everything is correct. I will scroll down to find
    and click the 'Confirm & Pay Deposit' button to complete the
    booking.
  Action:      Scroll down and click 'Confirm & Pay Deposit'.

Step 23
  Observation: I can see the 'Confirm & Pay Deposit' button at
    index 7. The total cost is '$11,440.00'.
  Reasoning:   I will click the button to finalize the booking and
    get the confirmation code.
  Action:      Click 'Confirm & Pay Deposit' button at index 7.

Step 24  [TASK COMPLETE]
  Observation: Booking confirmed.
    Confirmation Code: GEG-2026-05841
    Total Cost: $11,440.00
    Date: 2026-05-16
  Reasoning:   I have successfully completed all steps. The date
    May 16, 2026, was chosen as the only Saturday that satisfies
    both the budget (Yellow Tier) and the peak bloom for White
    Roses (May 13-19). Total = $3,200 + $7,200 + $1,040 =
    $11,440.00. The confirmation code and total cost match the
    ground truth exactly.
  Action:      Terminate with success.
\end{verbatim}
\end{validationbox}

\subsubsection{Analysis}

The 24-step trace reveals several important properties:

\noindent\textbf{Noise resilience.} The stochastic ``Schedule a Private Tour'' popup appeared at Step~2 and blocked interaction. The agent correctly dismissed it and resumed the search (Steps~2--3). The cookie consent banner was also encountered but did not block the workflow. This confirms that the Refinement Agent's injected noise elements are \emph{dismissible} and do not render the task unsolvable.

\noindent\textbf{Error recovery under real browser constraints.} The most revealing segment is Steps~12--19, where the agent struggled with the HTML date input format. The first attempt produced ``0516-02-02'' instead of the intended date, causing the venue rental to display as \$0. The agent required \textbf{8 attempts} across 12 steps to correctly input the date, trying ISO format (\texttt{YYYY-MM-DD}), US format (\texttt{MM/DD/YYYY}), direct URL navigation, and finally discovering that clicking another field after input was necessary to trigger the \texttt{onchange} event. This self-correction behavior demonstrates that the task tests not only knowledge retrieval but also \emph{robust interaction with real browser form controls}---a challenge absent from API-based benchmarks.

\noindent\textbf{Visual reasoning validity.} Steps~6--8 show the agent correctly extracting pricing tiers from the heatmap, identifying White Rose peak bloom from the flora chart, computing the set intersection (May 15--19), and selecting the unique Saturday (May 16). This validates that the programmatically generated charts contain sufficient visual information for reasoning.

\subsubsection{Validation Outcomes}

The Validation Agent's execution confirms several properties of the generated environment:
\begin{itemize}[nosep]
\item \textbf{Task solvability}: The correct answer is reachable through standard browser interactions (24 steps, no impossible actions required).
\item \textbf{Noise resilience}: The popup and cookie banner were encountered and dismissed without blocking the task.
\item \textbf{Form interaction fidelity}: HTML date inputs required correct format handling, reflecting real-world browser behavior rather than idealized APIs.
\item \textbf{Visual reasoning validity}: The heatmap and bloom chart contained sufficient information for the agent to derive the correct date intersection.
\item \textbf{Anti-cheating effectiveness}: The agent obtained the confirmation code only after completing the full booking workflow with correct parameters.
\end{itemize}

\noindent Tasks that fail validation are excluded from the final benchmark. This includes not only cases where the Validation Agent cannot reach the correct answer within the step limit, but also tasks with ground-truth mismatches, reasoning-logic flaws, repeated action failures, or rendering/runtime issues that block successful completion (e.g., broken interactions, missing elements after rendering, or subtle JavaScript execution bugs). As reported in Appendix~\ref{app:domain_dist}, the overall pipeline pass rate is 74.1\%, meaning 25.9\% of generated tasks are filtered out by this solvability check.

\end{document}